\PassOptionsToPackage{table,dvipsnames}{xcolor}
\documentclass[10pt,letterpaper]{segowam}
\pdfmapfile{=fonts/inter/Inter.map}
\usepackage{lmodern}

\usepackage{amssymb,mathtools}
\usepackage{float,tabularx,array,colortbl,xspace,wrapfig,needspace}
\usepackage{marvosym}
\definecolor{ourshade}{HTML}{E7F0F8}

\newcommand{\ourmethod}{SeGo-WAM\xspace}
\newcommand{\etal}{\emph{et al.}\xspace}
\newcommand{\corrmark}{*}
\newcommand{\embyes}{\ensuremath{\checkmark}}
\newcommand{\embno}{\ensuremath{\times}}
\newcommand{\embpartial}{\ensuremath{\sqrt{}\mkern-9mu{\smallsetminus}}}

\newtheorem{proposition}{Proposition}

\setcitestyle{authoryear,round,semicolon}
\hypersetup{pdftitle={SeGo-WAM: Does Action Need Appearance? Semantics and Geometry Suffice},pdfauthor={Yu Liu and Jialiang Zhang and Haitao Lin and Ziyi Cai and Yanqi Bao and Zhitong Huang and He Zhang and Shuo Wang and Zipei Fan and Jingbo Zhang}}
\renewcommand{\titlelogoraise}{-2mm}
\AtBeginDocument{\fancyhead{}
  \fancyhead[C]{\headerlogo{\raisebox{\titlelogoraise}{\begingroup%
  \makebox[\textwidth][l]{%
    \raisebox{-5mm}[13mm][0pt]{%
      \includegraphics[width=13mm,height=13mm,keepaspectratio]{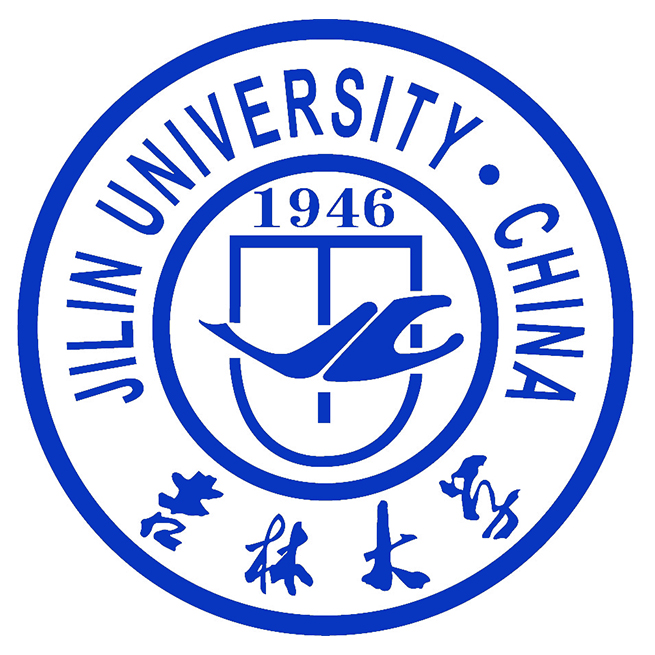}}%
    \hspace{3mm}%
    \raisebox{-5mm}[13mm][0pt]{%
      \includegraphics[width=13mm,height=13mm,keepaspectratio]{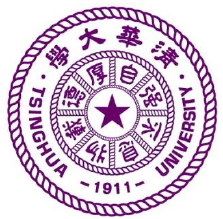}}%
    \hfill
    \raisebox{-1mm}[6.5mm][0pt]{%
      \includegraphics[width=29mm,height=6.5mm,keepaspectratio]{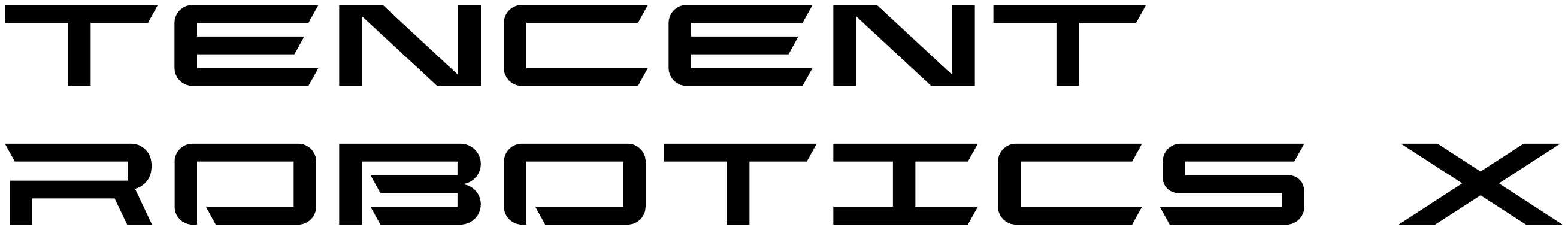}}%
    \hspace{6mm}%
    \raisebox{-3mm}[9mm][0pt]{%
      \includegraphics[width=24mm,height=9mm,keepaspectratio]{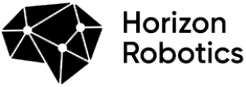}}%
    \hspace{5mm}%
    \raisebox{-2mm}[7.5mm][0pt]{%
      \includegraphics[width=21mm,height=7.5mm,keepaspectratio]{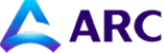}}%
  }%
\endgroup%
}}}}
\AtBeginDocument{\urlstyle{sf}}
\title{SeGo-WAM: Does Action Need Appearance?\\ Semantics and Geometry Suffice}
\author[1,2]{Yu~Liu}
\author[1]{Jialiang~Zhang}
\author[2,\dagger]{Haitao~Lin}
\author[3]{Ziyi~Cai}
\author[4]{Yanqi~Bao}
\author[5]{Zhitong~Huang}
\author[2]{He~Zhang}
\author[3]{Shuo~Wang}
\author[1,\corrmark]{Zipei~Fan}
\author[2,\dagger,\corrmark]{Jingbo~Zhang}
\affiliation[1]{Jilin University}
\affiliation[2]{Tencent Robotics X}
\affiliation[3]{Tsinghua University}
\affiliation[4]{Horizon Robotics}
\affiliation[5]{Tencent ARC Lab}
\contribution[\dagger]{Project Lead}
\contribution[\corrmark]{Corresponding Author}
\abstract{World action models (WAMs) typically condition actions on appearance-dominated RGB features, conflating what the action reads with what the world model predicts. We ask whether action needs appearance. Inspired by the what and where pathways of primate vision, we hypothesize that action requires only current and predicted semantics and geometry, with appearance serving as a learning signal. We present \ourmethod, which represents appearance, semantics and geometry as RGB, DINOv3 and metric-depth latents in a shared video diffusion transformer with a dense, modality-routed mixture of experts. A visibility mask restricts the action expert to semantic and geometric streams, so appearance prediction serves only for training and is omitted at deployment without changing actions. Initialized from a pretrained video generator, \ourmethod is trained in a single stage without embodied action pretraining. Ablations show that exposing appearance to the action or removing appearance prediction degrades randomized-scene success. \ourmethod achieves 91.04\% Easy and 55.20\% Hard success on RoboTwin clean-to-random transfer, outperforming methods with additional pretraining, and reaches 99.3\% on LIBERO and 79.70\% (trial-weighted) on LIBERO-Plus. These results suggest that action does not need appearance: semantics and geometry suffice.
We will make our code and model available upon acceptance.
}
\metadata[Project Website]{\url{https://rainyrobo.github.io/SeGo-WAM/}}
\metadata[Correspondence]{\href{mailto:eckertzhang@tencent.com}{\nolinkurl{eckertzhang@tencent.com}}\hfill\mbox{\textbf{Date:} November 25, 2026}}

\begin{document}
\suppressfloats[t]
\maketitle
\begin{figure}[t]
  \centering
  \includegraphics[width=\textwidth]{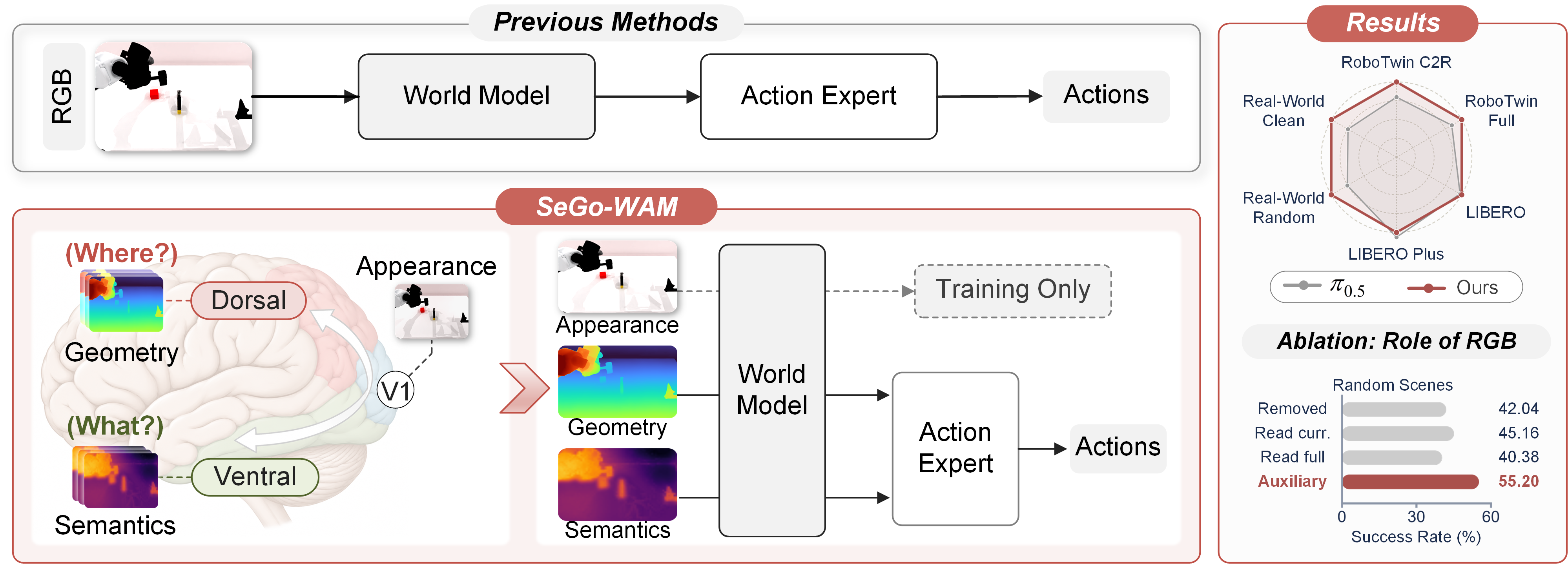}
  \caption{\textbf{\ourmethod learns from appearance but acts on semantics and geometry.} \textbf{Top left}: in most WAMs, the action expert reads appearance-dominated video latents. \textbf{Bottom left}: the ventral (``what'') and dorsal (``where'') pathways of primate vision, mapped to semantics and geometry. \textbf{Bottom center}: \ourmethod uses appearance only as training supervision and conditions actions on the semantic and geometric streams. \textbf{Right}: success rates on RoboTwin-C2R, RoboTwin-Full, LIBERO and LIBERO-Plus; real-world phone-packing progress (top) and RGB-role ablations on RoboTwin-C2R Hard (bottom).}
  \label{fig:overview}
\end{figure}

\section{Introduction}
\label{sec:intro}
World action models (WAMs) have rapidly become a leading paradigm for robot manipulation~\citep{wamsurvey2026,pad2024,uva2025,uwm2025,imagewam2026,maskwm2026}. By coupling a pretrained generative world model with an action expert, they jointly predict future observations and actions, inheriting rich spatiotemporal priors and achieving strong results in simulation and on real robots~\citep{fastwam2026,openwam_pku2026,prismwam2026,flexpi2026,imagewam2026,yu2026maskwam,xwam2026,fourdwam2026}. Much recent progress, however, comes from additional pretraining, such as embodied action pretraining~\citep{openwam_pku2026} or an extra world-pretraining stage~\citep{prismwam2026}. A more basic design choice has received far less scrutiny: which world features the action expert should read.

In most current WAMs, the answer is appearance: their world models build on video latents learned for pixel reconstruction, so the features the action expert reads are dominated by how the scene looks (Fig.~\ref{fig:overview}, top left); figuratively, the motor system reads the retina. Here, \emph{appearance} denotes photometric properties such as texture, illumination and background, rather than what the scene contains and where things are; it can change drastically while the task stays the same. Appearance shifts thus propagate into control: RGB-centric WAMs can lose most of their clean-scene success under scene randomization (Tab.~\ref{tab:comparison-robotwin}). Appearance in fact plays two roles that existing designs couple: it shapes what the action expert reads and is the target that the world model predicts. Methods that add geometry, semantics or masks~\citep{flexpi2026,prismwam2026,xwam2026,fourdwam2026,yu2026maskwam} or replace video generation with image editing~\citep{imagewam2026} keep both roles, whereas MWM~\citep{maskwm2026} keeps only the first. Since none of these designs separates the two roles, it remains unclear which of them action needs.
This raises a simple question: \emph{does action need appearance?} More precisely, must the action expert attend to appearance, and must the world model still predict it during training?

Primate vision suggests an answer (Fig.~\ref{fig:overview}, bottom left). The two-stream hypothesis separates a ventral pathway for \emph{what} an object is from a dorsal pathway for \emph{where} it is~\citep{ungerleider1982two}, a distinction later recast as vision for perception versus vision for action~\citep{goodale1992separate}. Both pathways start from the same retinal input, yet the motor system does not act on the image itself: the ventral stream selects the goal object, and the dorsal stream turns its spatial structure into reaching and grasping~\citep{milner2008two}. In manipulation, semantics plays the ventral role and geometry the dorsal one; that the target is the red cup is semantic, whereas the pixel color it takes under the current lighting is appearance. We therefore hypothesize that (H1) action needs only current and predicted semantics and geometry, and that (H2) appearance remains a valuable learning signal, since predicting it preserves the pretraining task of the inherited video generator and allows adaptation in a single training stage.

We instantiate these hypotheses as \ourmethod, which \emph{sees} through a semantic stream and \emph{goes} through a geometric stream (Fig.~\ref{fig:overview}, bottom center). \ourmethod represents appearance, semantics and geometry as RGB latents, compressed DINOv3 features~\citep{dinov3_2025} and metric-depth latents, modeled as three streams of a shared video DiT whose dense, modality-routed mixture of experts provides modality-specific capacity. A separate action DiT serves as the action expert and is coupled to these streams through joint attention~\citep{mot2024}. A visibility mask isolates the appearance stream: semantics and geometry read each other, the action expert reads their current and predicted future tokens, and no retained stream reads RGB latents. Appearance prediction therefore shapes the shared backbone only through training gradients, and the entire RGB generation stream can be dropped at deployment without changing the predicted actions; raw RGB supplies the frozen semantic encoder and, when depth is estimated from RGB, the depth estimator. Both DiTs are initialized from Wan2.2-5B~\citep{wan22_2025} and trained jointly in a single stage; apart from the frozen depth and semantic VAEs, which were trained without action labels, \ourmethod uses no embodied action pretraining or additional world-model pretraining stage.

Controlled experiments support both hypotheses (Fig.~\ref{fig:overview}, bottom right; Sec.~\ref{sec:analysis}). Under scene randomization, granting the action expert access to current RGB lowers success from 55.20\% to 45.16\%, and reading both current and future RGB lowers it to 40.38\%, supporting H1; removing the RGB branch lowers it to 42.04\%, supporting H2. With single-stage world--action training, \ourmethod achieves 91.04\% Easy and 55.20\% Hard success on the 50-task RoboTwin clean-to-random benchmark~\citep{robotwin2_2025} (73.12\% on average; Fig.~\ref{fig:overview}, top right), outperforming PrismWAM~\citep{prismwam2026} and OpenWAM-$\alpha$~\citep{openwam_pku2026} despite their extra pretraining. It further reaches 94.06\% average success on RoboTwin-Full, 99.3\% on LIBERO and 79.70\% trial-weighted success on LIBERO-Plus~\citep{liberoplus2026} (Tabs.~\ref{tab:comparison-robotwin} and~\ref{tab:comparison-libero}).

In summary, our contributions are threefold:
\begin{itemize}
  \setlength\itemsep{0pt}
  \item We identify two coupled roles of appearance in WAMs, what the action expert reads and what the world model predicts, and, drawing on the two-stream hypothesis, formulate hypotheses that action needs only semantics and geometry while appearance serves as a learning signal.
  \item We propose \ourmethod, which isolates RGB appearance latents from the action expert within a shared video DiT, enabling single-stage training from a pretrained video generator and deployment without RGB generation.
  \item Controlled studies support both hypotheses; with single-stage world--action training, \ourmethod achieves the best clean-to-random transfer among compared methods on RoboTwin, along with strong results on LIBERO and LIBERO-Plus.
\end{itemize}

\section{Related Work}
\label{sec:related}

\paragraph{World action models.}
Video prediction has long informed robot control~\citep{liu2026plaw}. VPP~\citep{vpp2024} conditions a policy on predictive features of a video diffusion model, PAD~\citep{pad2024} jointly denoises future images and actions, and UVA~\citep{uva2025} and UWM~\citep{uwm2025} couple video and action diffusion within a unified model. Current WAMs couple a large pretrained video generator with an action expert, typically through mixture-of-transformers attention~\citep{mot2024}, and OpenWAM~\citep{openwam_pku2026} studies this design space through modular world--action pretraining. Fast-WAM~\citep{fastwam2026} and GigaWorld-Policy~\citep{gigaworldpolicy2026} use video prediction only as a training signal and skip future generation at inference, while ImageWAM~\citep{imagewam2026} builds on an image-editing model instead. In all of these models, the action reads latents learned for image or video reconstruction.

\paragraph{Structured world representations.}
Many works enrich WAMs with representations beyond appearance. For geometry, X-WAM~\citep{xwam2026} and SA-WAM~\citep{sawam2026} predict depth alongside RGB, 4D-WAM~\citep{fourdwam2026} aligns intermediate features with 3D trajectory fields, and PrismWAM~\citep{prismwam2026} rolls out depth under observed-RGB guidance and sparse RGB-goal supervision. For semantics, GeoSem-WAM~\citep{geosemwam2026} adds auxiliary geometric and semantic heads alongside RGB prediction, while MWM~\citep{maskwm2026} and MaskWAM~\citep{yu2026maskwam} predict semantic masks, the former in place of pixels. Flex-$\pi$~\citep{flexpi2026} jointly predicts RGB, pointmaps and DINO features with stream dropout, UniFlowWAM~\citep{uniflowwam2026} pairs future DINOv3 features with 3D scene flow, and DreamVLA~\citep{dreamvla2025} forecasts depth and semantic cues for a vision-language-action model. Feature-space approaches include V-JEPA~2~\citep{vjepa2_2025}, which predicts in a self-supervised embedding space and adapts it for planning, ReWAM~\citep{rewam2026}, which shapes compact DINOv3 world states through action gradients, and R$^2$-WAM~\citep{representation_recipe2026}, which compares external visual representations for WAMs. Motion offers a further abstraction, through optical flow in FlowWAM~\citep{flowwam2026} and end-effector motion in MoWAM~\citep{mowam2026}, and geometric observation encoders such as DP3~\citep{dp3_2024} and DeFM~\citep{defm2026} show the value of 3D structure for control~\citep{lin2026universal}. Most of these methods keep RGB-derived features in the action pathway, and those that avoid appearance latents also forgo appearance prediction. \ourmethod separates the two roles: appearance remains a prediction target on the shared video DiT, while only the semantic and geometric streams reach the action.

\paragraph{What and where pathways in robot learning.}
The two-stream hypothesis~\citep{ungerleider1982two,goodale1992separate} has also inspired robot learning architectures. Jang~\etal~\citep{jang2017semantic} decompose semantic grasping into a ventral stream that recognizes object classes and a dorsal stream that predicts grasp success from geometric relations, and CLIPort~\citep{cliport2021} combines a CLIP-based semantic pathway with a Transporter-based spatial pathway for language-conditioned manipulation. These designs separate \emph{what} and \emph{where} in the observation encoders of a policy. \ourmethod applies the same separation to the predicted futures of a world action model and, rather than discarding appearance, retains it as a training signal on the shared backbone.

\section{Method}
\label{sec:arch}

\begin{figure}[t]
  \centering
  \includegraphics[width=\textwidth]{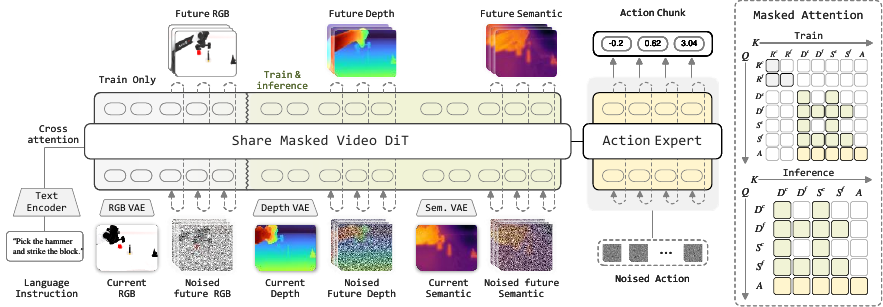}
  \caption{\textbf{\ourmethod architecture.} \textbf{Left}: RGB, depth and semantic streams in the shared video DiT, coupled to the action DiT; RGB generation is skipped at inference. \textbf{Right}: training and inference attention masks, with queries as rows, keys as columns and permitted attention as filled cells. $R$, $D$, $S$ and $A$ denote RGB, depth, semantic and action tokens; superscripts $c$ and $f$ denote current and future.}
  \label{fig:arch}
\end{figure}

\ourmethod keeps appearance as a prediction target but removes it from the action pathway. We formalize the two roles of appearance (Sec.~\ref{sec:setup}), then describe what the action reads instead (Sec.~\ref{sec:layouts}), how appearance still helps (Sec.~\ref{sec:loss}) and why RGB generation can be dropped at deployment (Sec.~\ref{sec:execution}).

\subsection{Problem Setup}
\label{sec:setup}
Given the current RGB observation $o_0$, language instruction $l$ and proprioceptive state $q_0$, a WAM predicts an action chunk $a=a_{0:H-1}$ jointly with the future evolution of the scene. A video DiT models appearance through RGB latents, denoising future latents $R^f$ conditioned on the clean current latent $R^c=\mathcal E_R(o_0)$ of a frozen VAE encoder $\mathcal E_R$, and an action DiT denoises the action tokens $A$; the two are coupled by joint attention~\citep{mot2024}.

Appearance can then affect the action in two ways: through \emph{access}, when the action attends to RGB tokens directly or through another stream, and through \emph{supervision}, when the world model is trained to predict RGB. \ourmethod keeps supervision but removes access.

Without access, the action must observe the scene through other streams. \ourmethod (Fig.~\ref{fig:arch}, left) therefore adds a depth stream and a semantic stream to the world model, whose current latents $D^c=\mathcal E_D(d_0)$ and $S^c=\mathcal E_S(s_0)$ encode the metric depth $d_0$ and the DINOv3 features $s_0=\Phi(o_0)$~\citep{dinov3_2025}, and whose future latents are $D^f$ and $S^f$. An action chunk is sampled from its policy as
\begin{equation}
  a=G_\theta(S^c,D^c,l,q_0;\epsilon),\qquad a\sim\pi_\theta(\cdot\mid s_0,d_0,l,q_0),
  \label{eq:policy}
\end{equation}
where $\epsilon$ denotes the initial Gaussian noise for the depth, semantic and action streams, $G_\theta$ jointly denoises $D^f$, $S^f$ and $A$ and returns $a$, and $\theta$ are the trainable parameters. The action thus depends on $o_0$ only through $s_0$, and through $d_0$ if depth is estimated from RGB.
Sec.~\ref{sec:layouts} specifies what the action expert attends to, and Sec.~\ref{sec:loss} how RGB prediction improves the policy without attention to RGB tokens.

\subsection{Acting on What and Where}
\label{sec:layouts}
Following the two-stream view (Sec.~\ref{sec:intro}), the action reads semantics for \emph{what} and geometry for \emph{where}. Semantic latents are obtained by compressing the 768-channel DINOv3 ViT-B/16 features of each view into 48-channel posterior means with a frozen semantic VAE (S-VAE) $\mathcal E_S$ and carry object identity and instruction-relevant attributes. Depth latents are obtained by encoding metric depth with a frozen 48-channel depth VAE $\mathcal E_D$~\citep{prismwam2026} and capture spatial relations, approach and contact.

An attention mask (Fig.~\ref{fig:arch}, right) restricts the action to these two streams and specifies how all streams interact. Writing $\mathcal V(X)$ for the tokens visible to $X$ at the modality and view level, with $R=R^c\cup R^f$ and likewise for $D$ and $S$, and $S_{\mathrm{head}}$ for current and future head-view semantic tokens, we define
\begin{equation}
  \mathcal V(R)=R,\qquad
  \mathcal V(D\cup S)=D\cup S,\qquad
  \mathcal V(A)=D\cup S_{\mathrm{head}}\cup A,
  \label{eq:mask}
\end{equation}
with the additional temporal restriction that current tokens attend only to current tokens. Under this mask, RGB tokens attend only to RGB tokens, and no other stream attends to them. Thus, RGB latents cannot influence action generation through cross-stream attention. Depth and semantics attend to each other, so the identity of the target and its spatial configuration constrain each other during prediction. The action attends to current and predicted depth and semantics, since control depends on how the scene will evolve, whereas no world token attends to the action, so the predicted world does not depend on the candidate action. Together, these constraints implement H1: action needs only current and predicted semantics and geometry.

The \emph{what}/\emph{where} division also selects the camera views. The head view covers the whole workspace and identifies the object the instruction refers to, whereas the wrist views capture the gripper and its surroundings at close range, providing local geometry for approach and contact. The action therefore reads semantics from the head view only and depth from all views. It remains to explain how the RGB stream, which no other stream attends to, can still improve the action.

\subsection{Appearance as a Learning Signal}
\label{sec:loss}
Although no other stream attends to RGB tokens, RGB prediction can still shape the depth and semantic streams, because all world streams share the transformer blocks of the video DiT. These blocks are initialized from the video generator Wan2.2-TI2V-5B~\citep{wan22_2025}, and each stream keeps only its own patch, text and time embeddings, time projection and output head. Since all three VAEs produce 48-channel latents, the depth and semantic interfaces are initialized as copies of the Wan interfaces but keep separate parameters. Predicting future RGB latents in the frozen Wan2.2 VAE space is exactly the task the shared blocks were pretrained on, so the RGB stream keeps training them on this task while they learn the new depth and semantic latents.

Sharing all blocks, however, forces a single dense FFN to fit three latent distributions. We therefore replace the FFNs of three late blocks with a dense, modality-routed mixture of experts (MoE)~\citep{moe2017}. For a token of stream $m$ with hidden state $h$ in layer $\ell$,
\begin{equation}
  \mathrm{FFN}^{\ell}_m(h)=\beta_\ell F^{\ell}_0(h)+(1-\beta_\ell)\sum_{k=1}^{K}g^{\ell}_k(m)\,F^{\ell}_k(h),
  \label{eq:moe}
\end{equation}
where $F^{\ell}_0$ is a shared expert, $F^{\ell}_{1:K}$ are $K$ routed experts, $\beta_\ell\in(0,1)$ is learned, and the gates $g^{\ell}(m)=\operatorname{softmax}(W_\ell e_m+b_\ell)$ depend only on a learned modality embedding $e_m$. All experts run on every token. The experts are upcycled from the pretrained FFN and $W_\ell,b_\ell$ are zero-initialized, so each MoE layer initially approximates the dense FFN (details in Appendix~\ref{supp:params}).

Both video and action DiTs are trained jointly in a single stage, without embodied action pretraining or an extra world-model pretraining stage. Following Fast-WAM~\citep{fastwam2026}, the action DiT is initialized by linearly interpolating the same Wan2.2 weights to a smaller width; it keeps dense FFNs, while the three VAEs, DINOv3 and the text encoder stay frozen. All streams share a flow-matching objective~\citep{flowmatching2022}: for stream $m\in\{R,D,S,A\}$ with clean target $z^m_0$ (future latents, or $a$ for the action) and Gaussian noise $\epsilon_m$, we form $z^m_t=(1-\sigma(t))\,z^m_0+\sigma(t)\,\epsilon_m$ with velocity target $u_m=\epsilon_m-z^m_0$, where $\sigma$ is the shifted schedule of Wan2.2. All world streams are noised at one sampled time $t_W$ with independent noise, which each stream encodes with its own time embedding; the action uses its own time $t_A$, and condition frames are not flow-noised. With $f^m_\theta$ the velocity prediction from all clean and noisy latents $\mathbf z$, conditioned on $l$ and $q_0$, the objective is
\begin{align}
  \mathcal L_m&=\mathbb E\big[w(t_m)\operatorname{MSE}_{\Omega_m}\big(f^m_\theta(\mathbf z,t_W,t_A),u_m\big)\big],\nonumber\\
  \mathcal L&=\lambda_A\mathcal L_A+\lambda_D\mathcal L_D+\lambda_S\mathcal L_S+\lambda_R\mathcal L_R+\mathcal L_{\rm router},
  \label{eq:loss}
\end{align}
where $t_m=t_A$ for $m=A$ and $t_W$ otherwise, $w$ is the scheduler weight, $\operatorname{MSE}_{\Omega_m}$ averages over supervised positions $\Omega_m$ (excluding conditions and padding), $\mathcal L_{\rm router}$ is a Switch-style load-balancing loss on the gates~\citep{fedus2022switch}, and all $\lambda$ are one.

Within this objective, the mask of Eq.~\eqref{eq:mask} confines the influence of the RGB loss. Because RGB tokens attend only to RGB tokens, $\mathcal L_R$ is independent of the action DiT parameters $\theta_A$ and of the activations $h_X$ of the depth, semantic and action tokens:
\begin{equation}
  \nabla_{\theta_A}\mathcal L_R=0,\quad
  \partial\mathcal L_R/\partial h_X=0\quad\forall X\in\{D,S,A\}.
  \label{eq:gradient}
\end{equation}
RGB supervision updates the shared world-model parameters, thereby shaping the depth and semantic representations used by the action DiT. Appearance thus shapes the representations that the action reads without being read itself, which is the mechanism behind H2: appearance remains a valuable learning signal.

\subsection{Deployment without RGB Generation}
\label{sec:execution}
Since no other stream attends to RGB tokens (Eq.~\eqref{eq:mask}) and the MoE gates depend only on modality (Eq.~\eqref{eq:moe}), the depth, semantic and action streams receive identical inputs at every layer and denoising step regardless of whether RGB tokens are present. Thus, with identical initial noise, removing the entire RGB stream leaves the predicted action unchanged; Appendix~\ref{supp:omission} provides a formal argument.

At deployment, we omit the RGB prediction stream and its VAE, while retaining RGB observations as inputs for semantic extraction and depth estimation. Specifically, DINOv3 and the S-VAE encode the current RGB observation into $S^c$, while the depth VAE encodes current depth into $D^c$. Depth can be sensed or estimated from RGB: our simulated benchmarks use ground-truth depth, whereas real-world experiments use estimated depth (Sec.~\ref{sec:exp-setup}). The model then jointly denoises $D^f$, $S^f$ and $A$, executing the resulting action chunk for a fixed horizon before replanning. This eliminates RGB-specific appearance encoding, denoising and decoding at inference while preserving the benefits of RGB prediction during training.

\section{Experiments}
\label{sec:exp}

\subsection{Experimental Setup}
\label{sec:exp-setup}
\paragraph{Benchmarks.}
We evaluate on RoboTwin~2.0~\citep{robotwin2_2025} and LIBERO~\citep{libero2023}, and use LIBERO-Plus~\citep{liberoplus2026} for explicit distribution-shift evaluation. RoboTwin contains 50 bimanual tasks evaluated under two scene conditions with 100 episodes per task each; following the official naming, Easy denotes clean scenes and Hard denotes randomized scenes. We report two RoboTwin training protocols (Tab.~\ref{tab:comparison-robotwin}). RoboTwin-C2R trains on clean demonstrations only; its frozen VAEs are pretrained on clean and randomized observations without actions (Appendix~\ref{supp:repr}). Easy is in distribution and Hard measures clean-to-randomized generalization. RoboTwin-Full trains on both clean and randomized demonstrations, so both conditions are in distribution. LIBERO has four ten-task suites (Spatial, Object, Goal and Long), with 50 trials per task. LIBERO-Plus has seven perturbation dimensions and 10,030 trials; its evaluation uses the LIBERO-trained checkpoint without Plus-specific fine-tuning.

\paragraph{Baselines.}
Tabs.~\ref{tab:comparison-robotwin} and~\ref{tab:comparison-libero} distinguish three training regimes. Large-scale embodied action pretraining is used by $\pi_{0.5}$~\citep{pi05_2025}, OpenWAM-$\alpha$~\citep{openwam_pku2026} and X-WAM~\citep{xwam2026}, which reports more than 5,800 hours of cross-embodiment pretraining data. PrismWAM~\citep{prismwam2026} uses depth adaptation followed by joint training, without embodied action pretraining. starVLA~\citep{starvla2026}, FastWAM~\citep{fastwam2026}, Spatial-Forcing~\citep{spatialforcing2025}, 4D-WAM~\citep{fourdwam2026} and \ourmethod use single-stage task-specific training without embodied action pretraining; here, single-stage refers to joint world--action training without an additional world-model pretraining stage. An independent dagger marks X-WAM, Spatial-Forcing, 4D-WAM, PrismWAM and \ourmethod for additional spatial information. Baseline numbers are taken from published results under each method's evaluation setting rather than rerun under a shared training budget; Appendix~\ref{supp:comparison} lists the sources.

\paragraph{Implementation and evaluation protocol.} Depth sources differ between settings. In simulation, the simulator provides ground-truth depth for both current observations and future targets. In the real world, we do not use sensor depth: a frozen Video Depth Anything model~\citep{video_depth_anything} estimates depth from RGB, emulating deployments where reliable depth sensing is unavailable (Sec.~\ref{sec:realworld}). At inference, the RGB generation stream is omitted (Sec.~\ref{sec:execution}), and the policy uses ten denoising steps and executes 32 actions per chunk (10 on LIBERO). Each reported \ourmethod result comes from a single evaluation run. Appendix~\ref{supp:eval} lists the per-benchmark training and inference settings; Appendices~\ref{supp:params} and~\ref{supp:repr} detail the architecture, camera layout, augmentation and preprocessing.

\begin{table}[t]
  \centering
  \footnotesize
  \caption{\textbf{Success rates (\%) on RoboTwin-C2R and RoboTwin-Full.} Shaded columns are Easy/Hard averages. \textbf{Bold} and \underline{underline} mark the best and second-best results.}
  \label{tab:comparison-robotwin}
  \setlength{\tabcolsep}{2.5pt}
  \renewcommand{\arraystretch}{1}
  \begin{tabular*}{\columnwidth}{@{\extracolsep{\fill}}lc>{\columncolor{black!6}}ccc@{\hspace{10pt}}>{\columncolor{black!6}}ccc@{}}
    \toprule
    \multirow{2}{*}{Method} & \multirow{2}{*}{\shortstack{Embodied\\Pretraining}} & \multicolumn{3}{c}{RoboTwin-C2R} & \multicolumn{3}{c}{RoboTwin-Full} \\
    \cmidrule(lr){3-5}\cmidrule(lr){6-8}
    & & \multicolumn{1}{c}{Average} & Easy & Hard & \multicolumn{1}{c}{Average} & Easy & Hard \\
    \midrule
    $\pi_{0.5}$~\citep{pi05_2025} & \embyes & 58.35 & 70.70 & 46.00 & 79.75 & 82.74 & 76.76 \\
    OpenWAM-$\alpha$~\citep{openwam_pku2026} & \embyes & 69.05 & \underline{89.40} & 48.70 & \underline{93.60} & \underline{93.74} & \underline{93.46} \\
    X-WAM$^{\dagger}$~\citep{xwam2026} & \embyes & 47.90 & 70.00 & 25.80 & 90.25 & 89.80 & 90.70 \\
    PrismWAM$^{\dagger}$~\citep{prismwam2026} & \embpartial & \underline{70.57} & 87.98 & \underline{53.16} & -- & -- & -- \\
    \midrule
    starVLA~\citep{starvla2026} & \embno & 24.84 & 46.52 & 3.16 & 88.25 & 88.18 & 88.32 \\
    FastWAM~\citep{fastwam2026} & \embno & 39.85 & 77.80 & 1.90 & 91.83 & 91.88 & 91.78 \\
    Spatial-Forcing$^{\dagger}$~\citep{spatialforcing2025} & \embno & 51.97 & 77.20 & 26.74 & -- & -- & -- \\
    4D-WAM$^{\dagger}$~\citep{fourdwam2026} & \embno & 61.65 & 81.50 & 41.80 & 91.34 & 91.92 & 90.76 \\
    \midrule
    \textbf{SeGo-WAM (Ours)}$^{\dagger}$ & \embno & \textbf{73.12} & \textbf{91.04} & \textbf{55.20} & \textbf{94.06} & \textbf{94.54} & \textbf{93.58} \\
    \bottomrule
  \end{tabular*}
  \par\smallskip
  \begin{minipage}{\linewidth}
    \footnotesize\raggedright\textit{Note.} \embpartial{}: multi-stage adaptation without embodied action pretraining; $^{\dagger}$: additional spatial information.
  \end{minipage}
\end{table}

\begin{table*}[t]
  \centering
  \footnotesize
  \caption{\textbf{Success rates (\%) on LIBERO and LIBERO-Plus.} Shaded columns are averages, trial-weighted for LIBERO-Plus. \textbf{Bold} and \underline{underline} mark the best and second-best results.}
  \label{tab:comparison-libero}
  \setlength{\tabcolsep}{1.8pt}
  \renewcommand{\arraystretch}{1.15}
  \begin{tabular*}{\linewidth}{@{\extracolsep{\fill}}lc>{\columncolor{black!6}}ccccc@{\hspace{6pt}}>{\columncolor{black!6}}cccccccc@{}}
    \toprule
    \multirow{2}{*}{Method} & \multirow{2}{*}{\shortstack{Emb.\\PT.}} & \multicolumn{5}{c}{LIBERO} & \multicolumn{8}{c}{LIBERO-Plus} \\
    \cmidrule(lr){3-7}\cmidrule(lr){8-15}
    & & \multicolumn{1}{c}{Avg.} & Spatial & Object & Goal & Long & \multicolumn{1}{c}{Avg.} & Camera & Robot & Lang. & Light & Bg. & Noise & Layout \\
    \midrule
    $\pi_{0.5}$ & \embyes & 96.85 & 98.8 & 98.2 & 98.0 & 92.4 & \textbf{84.76} & \underline{71.23} & \textbf{76.06} & 85.82 & 97.72 & 96.19 & \underline{87.32} & \textbf{86.23}\\
    OpenWAM-$\alpha$ & \embyes & \underline{99.3} & \textbf{99.6} & 99.6 & \textbf{99.8} & \underline{98.2} & 68.79 & 33.21 & \underline{75.87} & 87.57 & 97.02 & 85.59 & 39.85 & 77.38\\
    X-WAM$^{\dagger}$ & \embyes & 96.60 & 96.8 & 98.6 & 95.4 & 95.6 & 74.80 & \textbf{72.92} & 46.32 & 88.48 & 93.61 & 88.29 & 62.90 & 80.79\\
    PrismWAM$^{\dagger}$ & \embpartial & \textbf{99.35} & \underline{99.4} & \textbf{100.0} & 98.2 & \textbf{99.8} & 79.44 & 60.60 & 67.61 & \textbf{92.13} & \textbf{99.65} & \textbf{99.54} & 67.46 & \underline{81.70}\\
    \midrule
    starVLA & \embno & 96.6 & 97.8 & 98.6 & 96.2 & 93.8 & 77.83 & 64.04 & 57.16 & 85.75 & 93.78 & 96.00 & 80.95 & 77.25\\
    FastWAM & \embno & 97.6 & 98.2 & \textbf{100.0} & 97.0 & 95.2 & 52.56 & 16.01 & 46.26 & 72.93 & 81.96 & 55.48 & 39.41 & 66.49\\
    Spatial-Forcing$^{\dagger}$ & \embno & 98.5 & \underline{99.4} & 99.6 & 98.8 & 96.0 & 71.10 & 56.85 & 37.94 & 84.39 & 95.36 & 91.82 & 65.90 & 79.02\\
    4D-WAM$^{\dagger}$ & \embno & 98.6 & 98.8 & 99.5 & 98.6 & 97.4 & 71.01 & 45.15 & 64.26 & \underline{90.63} & 94.29 & 57.71 & 69.08 & 79.21\\
    \midrule
    \textbf{SeGo-WAM (Ours)}$^{\dagger}$ & \embno & \underline{99.3} & \textbf{99.6} & \underline{99.8} & \underline{99.6} & \underline{98.2} & \underline{79.70} & 58.41 & 56.19 & 87.05 & \underline{98.95} & \underline{97.21} & \textbf{90.32} & 80.59\\
    \bottomrule
  \end{tabular*}
  \par\smallskip
  \begin{minipage}{\linewidth}
    \footnotesize\raggedright\textit{Note.} Emb.\ PT.: embodied action pretraining; \embpartial{}: multi-stage adaptation without embodied action pretraining; $^{\dagger}$: additional spatial information. Lang.: language; Bg.: background.
  \end{minipage}
\end{table*}

\subsection{Main Results}
\label{sec:main-results}
\paragraph{In-distribution performance.}
On RoboTwin-C2R Easy, \ourmethod achieves a 91.04\% success rate, the highest among the compared methods (Tab.~\ref{tab:comparison-robotwin}). Despite not using additional embodied action pretraining, it outperforms OpenWAM-$\alpha$, which benefits from embodied action pretraining, by 1.64 percentage points. It also surpasses PrismWAM, which employs depth adaptation followed by joint training, and 4D-WAM, which incorporates trajectory-field-based spatial supervision, by 3.06 and 9.54 percentage points, respectively. On RoboTwin-Full, whose training includes randomized scenes so that both settings are in distribution, \ourmethod achieves the best results in both Easy and Hard settings, with success rates of 94.54\% and 93.58\%, respectively, yielding an average of 94.06\%, 0.46 percentage points above OpenWAM-$\alpha$. On LIBERO, \ourmethod achieves 99.3\% success, demonstrating competitive performance with existing methods (Tab.~\ref{tab:comparison-libero}).

\paragraph{Distribution shift.}
On RoboTwin-C2R Hard, \ourmethod achieves 55.20\% success, 2.04 and 6.50 percentage points above PrismWAM and OpenWAM-$\alpha$, respectively (Tab.~\ref{tab:comparison-robotwin}). On LIBERO-Plus, it reaches 79.70\% overall success, ranking second among the compared methods and first among those without embodied action pretraining (Tab.~\ref{tab:comparison-libero}). Its advantage is strongest under appearance shifts: across lighting, background and sensor noise, it has the highest macro-average (95.49\%) and highest minimum success rate (90.32\%) among the compared methods. This pattern is consistent with the modality design: RGB prediction continues the pretrained video-generation task in the shared video DiT, while the attention mask prevents RGB tokens from reaching the action expert. The action instead reads semantic features for target identity and instruction-relevant attributes, and depth features for spatial relations relevant to approach and contact.

\Needspace{320pt}
\begin{wrapfigure}{r}{0.50\textwidth}
  \centering
  \includegraphics[width=\linewidth]{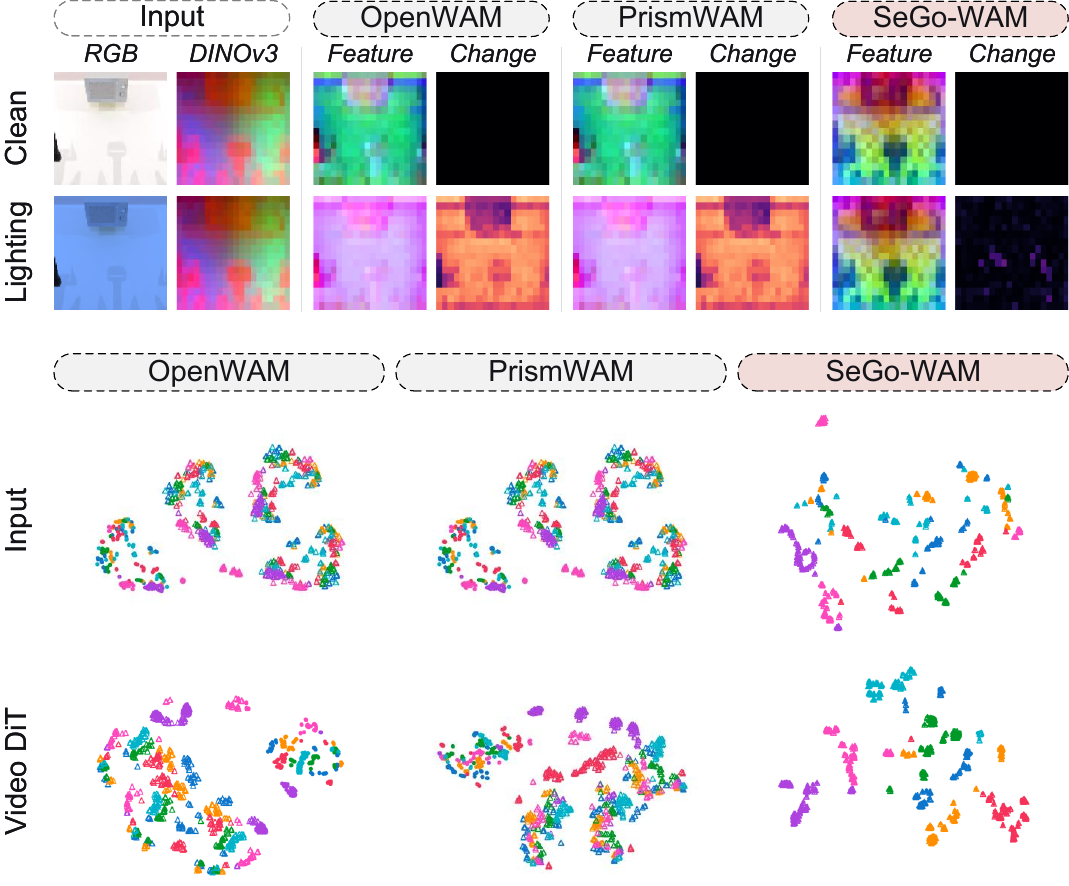}
  \caption{\textbf{Representations under RoboTwin lighting shifts.} \textbf{Top}: RGB inputs, DINOv3 features before S-VAE compression and each model's input encodings, with cosine distance to the clean features (brighter: larger change). \textbf{Bottom}: t-SNE of input encodings and final-layer RGB and semantic tokens over seven tasks. Colors denote tasks; filled circles and open triangles denote clean and perturbed inputs.}
  \label{fig:qualitative}
  \vspace{-15pt}
\end{wrapfigure}

\paragraph{Qualitative evaluation.}
Fig.~\ref{fig:qualitative} visualizes how RGB and semantic representations evolve under RoboTwin lighting perturbations, from input encoding to the final Video DiT layer. Specifically, we compare the RGB-VAE representations used by OpenWAM and PrismWAM with the DINOv3-based semantic representations used by \ourmethod. At the input level, RGB-VAE features exhibit pronounced changes under varying illumination, whereas DINOv3 features show greater stability. In the lower t-SNE panels, after processing by the Video DiT, \ourmethod's semantic features form clearer task-specific clusters, with samples from the same task remaining closely grouped across clean and perturbed conditions. In contrast, RGB-VAE features exhibit larger perturbation-induced shifts. The spatial feature maps at the top provide complementary evidence, showing greater structural consistency in semantic features than in RGB features; the separate DINOv3 column shows semantic features before S-VAE compression.

This behavior is consistent with \ourmethod's modality design. RGB future prediction continues to train the shared Video DiT on its pretrained video-generation objective, providing an auxiliary learning signal. Meanwhile, the attention mask prevents RGB tokens from being accessed by the depth and semantic streams, while allowing these two streams to interact in modeling object identity and spatial configuration. This enables the shared backbone to retain the benefits of video pretraining while developing task-relevant semantic and geometric representations. The action expert is conditioned only on current and predicted semantic and depth representations, reducing the direct propagation of appearance variations into action conditioning. Appendix~\ref{supp:qual} presents a complementary analysis under LIBERO-Plus fog and details the extraction protocol.

\Needspace{410pt}
\subsection{Analysis and Discussion}
\label{sec:analysis}

\begin{wraptable}{r}{0.50\textwidth}
  \centering
  \vspace{-10pt}
  \caption{\textbf{Modality-role ablations on RoboTwin-C2R (success, \%).} Each group varies one modality; parentheses indicate the reference setting. Avg.: Easy/Hard mean.}
  \label{tab:modality-roles}
  \begingroup
    \definecolor{modalitygroup}{HTML}{ECECEC}
    \footnotesize
    \setlength{\tabcolsep}{3pt}
    \begin{tabularx}{\linewidth}{@{}>{\raggedright\arraybackslash}X*{2}{>{\centering\arraybackslash}p{0.13\linewidth}}>{\centering\arraybackslash}p{0.13\linewidth}@{}}
      \toprule
      \textbf{Modality use} & \textbf{Easy} & \textbf{Hard} & \textbf{Avg.} \\
      \midrule
      \textbf{\ourmethod} & \textbf{91.04} & \textbf{55.20} & \textbf{73.12} \\
      \rowcolor{modalitygroup}\multicolumn{4}{@{}l}{\strut\textit{RGB} (auxiliary supervision)} \\
      Removed & 78.46 & 42.04 & 60.25 \\
      Read current & 85.24 & 45.16 & 65.20 \\
      Read current + future & 86.32 & 40.38 & 63.35 \\
      \rowcolor{modalitygroup}\multicolumn{4}{@{}l}{\strut\textit{Depth} (current + future)} \\
      Removed & 87.30 & 41.18 & 64.24 \\
      Auxiliary supervision & 87.54 & 43.34 & 65.44 \\
      Read current & 89.46 & 48.36 & 68.91 \\
      \rowcolor{modalitygroup}\multicolumn{4}{@{}l}{\strut\textit{Semantics} (current + future, head mask)} \\
      Removed & 80.12 & 43.10 & 61.61 \\
      Auxiliary supervision & 85.22 & 48.42 & 66.82 \\
      Read current & 87.68 & 50.94 & 69.31 \\
      Read current + future & 89.78 & 51.72 & 70.75 \\
      \bottomrule
    \end{tabularx}
  \endgroup
\end{wraptable}

\paragraph{RQ1: Should the action read all modalities predicted by the world model?}
The results distinguish the value of a modality for world-model supervision from its value as an action condition. In Tab.~\ref{tab:modality-roles}, Removed disables the modality stream and its prediction loss; Auxiliary supervision retains prediction without action access. Removing RGB prediction, while retaining RGB inputs to DINOv3, reduces success by 12.58 and 13.16 percentage points on Easy and Hard, respectively, demonstrating the value of appearance prediction as auxiliary supervision (H2). Conversely, allowing the action to read current RGB reduces success by 5.80 points on Easy and 10.04 points on Hard; exposing predicted RGB increases the Hard performance drop to 14.82 points. Thus, retaining RGB prediction while excluding it from action conditioning yields the strongest performance, supporting the separation of appearance modeling from action generation (H1).

\par
\ifnum\value{WF@wrappedlines}>1
  \vspace{\dimexpr\value{WF@wrappedlines}\baselineskip-3\baselineskip\relax}
\fi
\WFclear
\paragraph{RQ2: What are the respective contributions of depth and semantics?}
Depth and semantics provide complementary information for action generation. In Tab.~\ref{tab:modality-roles}, removing depth reduces success by only 3.74 points on Easy but 14.02 points on Hard, whereas removing semantics causes large drops in both settings (10.92 and 12.10 points). This pattern is consistent with geometry providing spatial information that matters most under scene variations, while semantics supports task-relevant object identification in both clean and randomized scenes, corresponding to their respective \emph{where} and \emph{what} roles (Sec.~\ref{sec:layouts}). Both modalities benefit from direct action conditioning beyond auxiliary prediction: reading current and predicted depth increases average success from 65.44\% to 73.12\%, while semantic readout increases it from 66.82\% to 70.75\%. Furthermore, restricting semantic action readout to head-view features (head mask) while retaining wrist-view prediction further improves average success to 73.12\% and Hard success from 51.72\% to 55.20\%. These results highlight the complementary roles of geometry and semantics and the importance of selective action conditioning.

\paragraph{RQ3: Does access to predicted futures improve action generation?}
The benefit of future conditioning depends on the modality (Tab.~\ref{tab:modality-roles}). Adding predicted future depth to the current observation improves Hard success from 48.36\% to 55.20\% (+6.84 percentage points), whereas future semantics provides a smaller gain of 0.78 points (50.94\% to 51.72\%). In contrast, additionally exposing predicted RGB reduces Hard success from 45.16\% to 40.38\% ($-4.78$ points). On Easy scenes, the corresponding changes remain within 2.10 points across all three modalities. These results highlight the value of predictive geometry under distribution shift, with semantic futures providing complementary information, while additional RGB conditioning reduces Hard success.

\subsection{Real-World Results}
\label{sec:realworld}
\begin{figure*}[t]
  \centering
  \includegraphics[width=\textwidth]{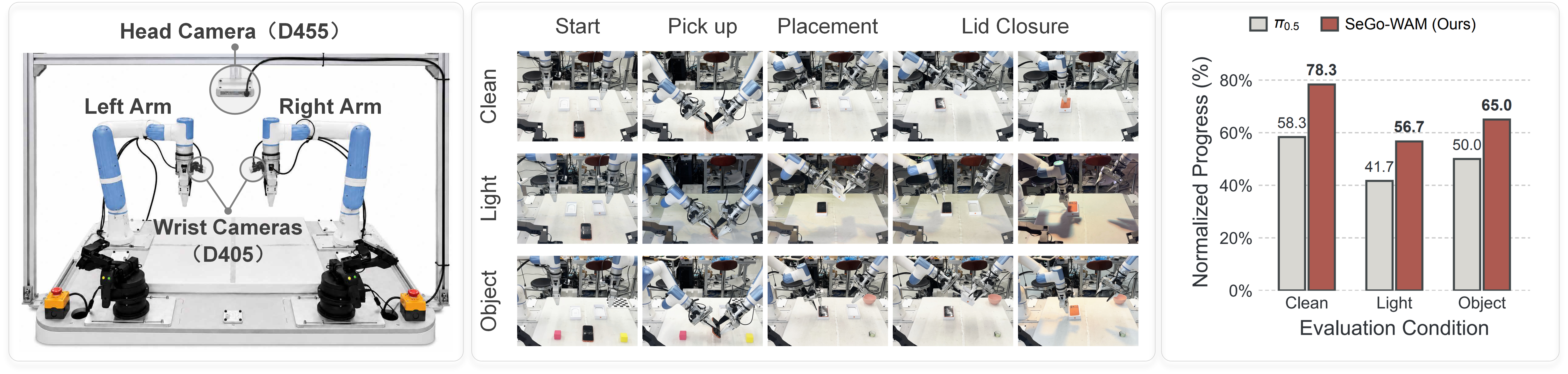}
  \caption{\textbf{Real-world phone packing.} \textbf{Left}: bimanual platform. \textbf{Middle}: executions under the Clean, Light and Object conditions (rows). \textbf{Right}: normalized progress (\%) over 20 rollouts per condition, scored on three sequential stages: pickup, placement and lid closing.}
  \label{fig:realworld-task}
\end{figure*}

\paragraph{Setup.}
We evaluate SeGo-WAM on a long-horizon phone-packing task using a bimanual robot, comprising three sequential stages: picking up a phone, placing it in a box, and closing the lid (Fig.~\ref{fig:realworld-task}). The model is trained on 305 teleoperated demonstrations collected under clean conditions, with a 95/5 training-validation split. Observations include one head camera, two wrist cameras, and end-effector poses with gripper commands. For depth input, we estimate depth from RGB using the frozen Video Depth Anything model and encode it with the frozen depth VAE, preserving the semantic-geometric interface used in simulation. We compare against $\pi_{0.5}$ fine-tuned on the same demonstrations. Evaluation comprises 20 rollouts in each of three conditions: Clean, Light (illumination changes), and Object (random distractors). We measure normalized progress by the fraction of completed stages across all rollouts, with each stage contributing one point. Further protocol details are provided in Appendix~\ref{supp:realworld}.

\paragraph{Performance and efficiency.}
Figure~\ref{fig:realworld-task} summarizes the results. Under the Clean condition, SeGo-WAM achieves 78.3\% normalized progress, outperforming $\pi_{0.5}$ (58.3\%) by 20.0 percentage points. The advantage persists under distribution shifts, reaching 56.7\% under Light and 65.0\% under Object, compared with 41.7\% and 50.0\% for $\pi_{0.5}$, respectively. Averaged across the two perturbed conditions, SeGo-WAM achieves 60.8\% progress, a 15.0-point improvement over $\pi_{0.5}$ (45.8\%). These results demonstrate the effectiveness of semantic-geometric action conditioning in real-world long-horizon manipulation, including under illumination changes and object clutter. Appendix~\ref{supp:realworld} additionally presents qualitative executions of bowl stacking and test-tube transfer under dynamic colored lighting and added distractors; these additional tasks are not included in the quantitative phone-packing evaluation.

In a separate single-GPU benchmark of deployment cost, \ourmethod takes a median of 574.2\,ms per 32-action chunk with its default preset, corresponding to an amortized 17.94\,ms per executed action, compared with 5.58\,ms for OpenWAM-$\alpha$ and 33.26\,ms for PrismWAM under their respective default presets. Appendix~\ref{supp:details} reports the measurement protocol and all configurations.

\section{Conclusion}
We introduce SeGo-WAM, a modality-decoupled world action model that retains future RGB prediction as an auxiliary objective on a shared video-pretrained backbone, while restricting action conditioning to current and predicted semantic and geometric latents through attention masking. SeGo-WAM achieves 91.04\% and 55.20\% success on RoboTwin-C2R Easy and Hard, respectively, leading the compared methods, and 79.70\% trial-weighted success on LIBERO-Plus. Ablations demonstrate that RGB prediction provides beneficial supervisory signals, whereas exposing RGB latents to the action expert degrades performance, particularly under distribution shift. These findings support decoupling predictive supervision from action conditioning, with semantic and geometric latents providing the task-relevant representations for action generation.

SeGo-WAM has two limitations. Encoding and predicting semantic and geometric latents adds inference overhead (Appendix~\ref{supp:details}), and VAE compression of semantic features may discard fine visual details needed for tasks with subtle appearance cues. Future work will reduce this overhead and develop semantic encoders that better balance robustness and detail preservation.

\FloatBarrier
\bibliographystyle{plainnat}
{\small\setlength{\bibsep}{3pt}\bibliography{main}}
\clearpage
\setcounter{figure}{0}
\setcounter{table}{0}
\setcounter{equation}{0}
\renewcommand{\thefigure}{S\arabic{figure}}
\renewcommand{\thetable}{S\arabic{table}}
\renewcommand{\theequation}{S\arabic{equation}}
\renewcommand{\theHfigure}{supp.\arabic{figure}}
\renewcommand{\theHtable}{supp.\arabic{table}}
\renewcommand{\theHequation}{supp.\arabic{equation}}
\suppressfloats[t]
{\huge\sffamily\color{accentcolor}Appendix\par}
\appendix
\section{Appendix Overview}
\label{supp:overview}
The appendix documents the implementation and evaluation of \ourmethod, including representation preprocessing, training, RGB-generation omission, ablations and additional experimental results. Tab.~\ref{tab:roadmap} links these sections to the corresponding analyses in the main text.

\begin{table}[H]
  \centering\footnotesize
  \caption{\textbf{Appendix sections and the main-text sections they support.}}
  \label{tab:roadmap}
  \setlength{\tabcolsep}{3pt}
  \begin{tabularx}{\textwidth}{@{}c>{\raggedright\arraybackslash}Xl@{}}
    \toprule
    App. & Content & Main paper \\
    \midrule
    \multicolumn{3}{@{}l}{\textit{Method details}} \\
    \ref{supp:repr} & Views, depth, semantics, frozen assets, augmentation & Sec.~\ref{sec:layouts} \\
    \ref{supp:params} & Initialization, attention, MoE, objective, routing & Sec.~\ref{sec:loss} \\
    \ref{supp:omission} & RGB-omission invariance and paired check & Sec.~\ref{sec:execution} \\
    \addlinespace[2pt]
    \multicolumn{3}{@{}l}{\textit{Experimental protocol}} \\
    \ref{supp:eval} & Benchmarks, training and inference settings & Sec.~\ref{sec:exp-setup} \\
    \ref{supp:comparison} & Baseline sources and observed modalities & Tabs.~\ref{tab:comparison-robotwin}, \ref{tab:comparison-libero} \\
    \addlinespace[2pt]
    \multicolumn{3}{@{}l}{\textit{Additional results}} \\
    \ref{supp:details} & Inference cost across configurations & Sec.~\ref{sec:realworld} \\
    \ref{supp:ablation} & Ablation recipe and MoE ablation & Tab.~\ref{tab:modality-roles} \\
    \ref{supp:qual} & Representation analysis, paired rollouts & Fig.~\ref{fig:qualitative} \\
    \ref{supp:realworld} & Real-world protocol, stage scores, extra-task rollouts & Fig.~\ref{fig:realworld-task} \\
    \ref{supp:pertask} & Per-task RoboTwin-C2R and Full results & Tab.~\ref{tab:comparison-robotwin} \\
    \bottomrule
  \end{tabularx}
\end{table}

\section{Representations and Preprocessing}
\label{supp:repr}
Frozen VAEs encode depth and semantic features for world modeling and action conditioning, while RGB latents serve as auxiliary prediction targets. The following specifications define the camera layouts, modality encodings, pretraining data and augmentations used to construct these inputs and targets.

\paragraph{Camera layout and temporal sampling.}
RoboTwin provides one head view and two wrist views, each at $240\times320$. These are resized and tiled into a $384\times320$ canvas, with $256\times320$ pixels for the head view and $128\times160$ for each wrist view. LIBERO and LIBERO-Plus place the main and wrist views side by side in a $224\times448$ canvas. Each sample contains 33 states and 32 actions. Sampling an image every fourth state yields nine visual frames. Each RoboTwin world stream contains 360 tokens, including 120 current-condition tokens.

\paragraph{Depth encoding.}
Simulation provides ground-truth depth for current conditions and future targets. RoboTwin depth is converted from millimeters to meters, clipped to 0--1.5\,m and mapped to $[-1,1]$ before encoding with the frozen PrismWAM depth VAE~\citep{prismwam2026}. The resulting 48-channel latents use depth-specific normalization statistics. Geometric augmentation uses nearest-neighbor interpolation on raw depth before resizing to the encoder's view layout.

\paragraph{Semantic encoding.}
Frozen DINOv3 ViT-B/16 extracts 768-channel patch features. The frozen S-VAE compresses these features to 48-channel posterior means, followed by parameter-free LayerNorm over channels ($\epsilon=10^{-6}$). These normalized means define the semantic conditions and future targets. The S-VAE is trained separately to reconstruct standardized DINO features using feature MSE, cosine and KL losses over valid elements. Normalization statistics are computed from valid training frames. Its encoder and decoder each contain three Transformer layers with width 768, 16 attention heads and FFN width 2048; the bottleneck preserves the spatiotemporal token grid. The semantic features can retain color and fine visual attributes.

\paragraph{Visual encoder pretraining.}
We reuse PrismWAM's pretrained depth VAE and train the S-VAE under the same data protocol (Tab.~\ref{tab:frozen-assets}). The RoboTwin S-VAE uses Flex-$\pi$ DINO features~\citep{flexpi2026}; depth-VAE pretraining uses approximately 2,300 hours of AgiBotWorld-Beta~\citep{agibotworld2025} and RoboTwin depth~\citep{prismwam2026}. Both VAEs are pretrained on clean and randomized RoboTwin observations using reconstruction objectives without action supervision, then frozen during joint world--action training. The LIBERO S-VAE is trained for 150,000 updates on the four standard suites, with a task-stratified split of 1,671 training and 40 validation episodes and no LIBERO-Plus perturbations. RoboTwin-C2R uses clean demonstrations for joint world--action training; its frozen VAEs are pretrained on both clean and randomized observations.

\begin{table}[H]
  \centering\footnotesize
  \caption{\textbf{Frozen visual encoders and VAEs.} The S-VAE and depth VAE are trained with reconstruction and KL losses.}
  \label{tab:frozen-assets}
  \setlength{\tabcolsep}{6pt}
  \begin{tabularx}{\textwidth}{@{}l>{\raggedright\arraybackslash}X>{\raggedright\arraybackslash}p{0.24\textwidth}c@{}}
    \toprule
    Component & Pretraining data & Domain exposure & Action supervision \\
    \midrule
    DINOv3 / Wan2.2 RGB VAE & Upstream image / video & Upstream corpora & No \\
    RoboTwin S-VAE & Flex-$\pi$ RoboTwin DINO features~\citep{flexpi2026} & RoboTwin clean + randomized & No \\
    LIBERO S-VAE & Standard LIBERO, four suites & No LIBERO-Plus perturbations & No \\
    PrismWAM depth VAE & AgiBotWorld-Beta + RoboTwin depth~\citep{prismwam2026} & RoboTwin clean + randomized & No \\
    \bottomrule
  \end{tabularx}
\end{table}

\paragraph{RGB encoding.}
The frozen Wan2.2 VAE defines the RGB latent space, using normalization statistics separate from those of the semantic VAE.

\paragraph{Augmentation.}
Each training clip is augmented with probability 0.5. Each camera then independently draws one of three groups: photometric, noise, or both. The photometric group applies color jitter (brightness, contrast and saturation 0.2; hue 0.03) in random order, followed by gamma correction with an exponent drawn from $[0.9, 1.1]$. The noise group adds Gaussian noise with $\sigma = 0.01$ to RGB values in $[0, 1]$. All groups also crop each view to a side-length ratio drawn from $[0.95, 1.0]$ at a fixed aspect ratio and rotate it within $\pm5^\circ$. All parameters and noise samples are drawn per camera and fixed across frames. Geometric transforms are shared by RGB and native depth, with bilinear interpolation for RGB and nearest-neighbor interpolation with zero fill for depth; RGB is then clipped to $[0, 1]$. Depth receives no photometric or noise augmentation. DINO features are extracted online from the augmented RGB images. Action labels remain unchanged, coarse dropout is disabled, and validation and inference use no augmentation. Temporal windows are sampled uniformly, and padded targets are excluded from the loss. All modality ablations use these augmentations.

\section{Architecture and Training Details}
\label{supp:params}
This section specifies the network initialization, attention, expert routing and optimization used to implement Eqs.~\eqref{eq:mask}--\eqref{eq:loss}. It also reports the learned routing distributions and model parameter counts.

\begin{figure}[H]
  \centering
  \begin{subfigure}[t]{0.435\linewidth}
    \centering
    \includegraphics[height=149pt]{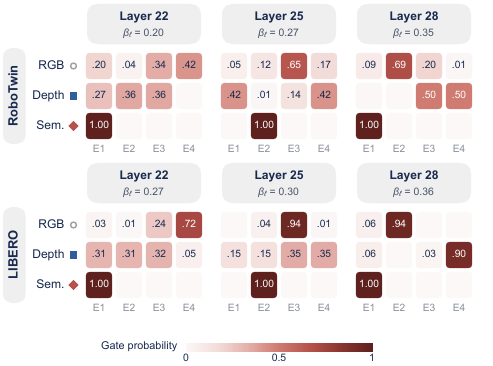}
    \caption{}
    \label{fig:router-gates}
  \end{subfigure}\hfill
  \begin{subfigure}[t]{0.555\linewidth}
    \centering
    \includegraphics[height=149pt]{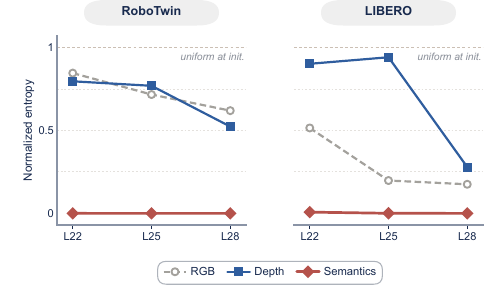}
    \caption{}
    \label{fig:router-entropy}
  \end{subfigure}
  \caption{\textbf{Modality routing in the MoE blocks.} (a)~Expert gate probabilities; panel titles give $\beta_\ell$, and blank cells denote values below 0.005. (b)~Gate entropy normalized by $\log 4$; dotted lines mark uniform gates.}
  \label{fig:router-weights}
\end{figure}

\paragraph{Parameters and initialization.}
Each world stream has separate patch, text and time embeddings, a time projection and an output head. The depth and semantic interfaces are initialized from Wan2.2 and updated independently. The world streams share Transformer blocks, including the query, key and value projections and MoE layers. The world tower has 30 layers of width 3072 and is initialized from Wan2.2-TI2V-5B. The action tower has 30 layers of width 1024, is initialized by linear interpolation of the same weights, and uses separate projections and dense FFNs. DINOv3, the three VAEs and the text encoder remain frozen.

\paragraph{Attention.}
Queries, keys and values are computed from the preceding layer in every stream. In addition to the modality mask in Eq.~\eqref{eq:mask}, current tokens are not flow-noised, are excluded from the loss and attend only to current tokens. Current RGB tokens attend to current RGB, while current depth and semantic tokens attend to each other. Future tokens use bidirectional temporal attention; world tokens do not attend to action tokens. At every semantic frame, the action expert reads semantic tokens only from head-view positions: the upper two thirds of the RoboTwin layout and the left half of the LIBERO layout. Wrist-view semantics remain prediction targets and can influence action indirectly through depth--semantic attention.

\paragraph{MoE.}
World-tower blocks 22, 25 and 28, numbered from one, use $K=4$ routed experts and one shared expert~\citep{dai2024deepseekmoe}; all experts process every token. The action tower uses dense FFNs. Following sparse upcycling~\citep{komatsuzaki2023sparse}, each routed FFN copies the pretrained dense FFN and receives a fixed-seed Gaussian weight perturbation with standard deviation 0.02 times that of the source tensor. The shared FFN is copied without perturbation. Its coefficient $\beta_\ell=\operatorname{sigmoid}(a_\ell)$ is learned per layer and shared across modalities, with $a_\ell=0$ at initialization ($\beta_\ell=0.5$). In Eq.~\eqref{eq:moe}, the gates are $g^{\ell}(m)=\operatorname{softmax}(W_\ell e_m+b_\ell)$, where $e_m$ is shared across MoE layers and $W_\ell,b_\ell$ are layer-specific. The gates depend only on modality and therefore remain constant across inputs for a trained checkpoint. The router loss follows Switch Transformers~\citep{fedus2022switch}:
\begin{equation}
  \mathcal L_{\rm router}=\lambda_{\rm r}\,K\sum_{k=1}^{K} f_k P_k,
\end{equation}
where $P_k$ is expert $k$'s mean gate weight across modalities and MoE layers, and $f_k$ is the fraction of modality--layer pairs in which it receives the largest gate; ties select the lower expert index. The shared expert is excluded. The router loss uses $\lambda_{\rm r}=0.01$, no entropy term and a linear warmup over the first 5\% of the initial training schedule: 1,611 updates for RoboTwin-C2R and 793 for LIBERO.

\paragraph{Objective.}
Each modality loss in Eq.~\eqref{eq:loss} averages squared errors over valid elements within each sample, applies the scheduler weight and then averages over samples. Image- and action-padding masks exclude padded targets, and current conditions are excluded from the loss. All four modality weights are one.

\paragraph{Optimization and noise schedules.}
Joint training uses AdamW with $\beta=(0.9,0.95)$ and BF16 mixed precision. The learning rate warms up linearly over the first 5\% of updates to $10^{-4}$, then decays to 1\% of this peak with a cosine schedule. LIBERO training is extended from 15 to 30 epochs by resuming the full training state after 15,870 updates. The update count is retained, and the schedule is adjusted to a total of 31,740 updates without restarting warmup. At resumption, the learning rate increases from $10^{-6}$ to approximately $5.46\times10^{-5}$, then decays to $10^{-6}$. The router-loss warmup remains 793 updates. Video and action flow schedules use 1,000 training timesteps, with timestep shift~5~\citep{sd32024} in both training and inference. For $u\sim\mathcal U(0,1)$, the shifted noise level is $\sigma=5u/(1+4u)$. As regularization, each current depth or semantic latent $c$ is replaced during training, with probability 0.1, by $(1-\alpha)c+\alpha\epsilon$, where $\alpha\sim\mathcal U(0,0.2)$ and $\epsilon$ is Gaussian noise with the mean and standard deviation of $c$. Current RGB latents and inference inputs are left unperturbed. Tab.~\ref{tab:train-config} lists the epochs and batch sizes for each benchmark.

\paragraph{Learned routing.}
Fig.~\ref{fig:router-weights} reports gate probabilities and normalized entropy for one training run per benchmark: RoboTwin-C2R at step 32,220 and LIBERO at step 31,740. Zero initialization of the router projections gives uniform gates and normalized entropy 1 for every modality. After training, the semantic stream concentrates on a single routed expert in every layer of both checkpoints (largest gate $\geq0.998$, normalized entropy $\leq0.008$), and this expert differs from the dominant RGB expert in all six layers. RGB and depth have more distributed gate weights (normalized entropy 0.18--0.94), and depth gates include exact ties between two experts in four of the six layers. Across the nine modality--layer pairs of each checkpoint, each routed expert receives the largest gate in at least one pair, indicating specialization across modalities. The shared-expert coefficient decreases from its initial 0.5 in every layer and grows with depth (0.20, 0.27 and 0.35 on RoboTwin-C2R; 0.27, 0.30 and 0.36 on LIBERO). The two checkpoints share the same fixed upcycling perturbations, and their dominant RGB (E4/E3/E2) and semantic (E1/E2/E1) experts coincide, whereas their depth assignments differ. Tab.~\ref{tab:moe-ablation} reports the routing ablation.

\paragraph{Parameter counts.}
The RoboTwin model contains approximately 7.258B parameters in the world--action network, depth/semantic interfaces and proprioceptive projection, excluding frozen encoders and codecs. Frozen components comprise T5 (5.681B), DINOv3 (85.697M), the RGB VAE (704.689M), depth VAE (704.599M) and S-VAE (33.200M).

\section{RGB-Generation Omission: Invariance and Verification}
\label{supp:omission}
We establish the conditions under which RGB-generation omission preserves action predictions and evaluate this property on a trained checkpoint. The analysis follows the attention mask and modality-dependent routing in Sec.~\ref{sec:execution}.

\paragraph{Retained computation.}
Deployment retains the semantic and depth encoders, their world streams and the action tower. It omits RGB VAE operations, RGB tokens, the RGB output head and RGB scheduler updates while keeping all model weights fixed. RGB observations continue to supply the DINOv3 encoder.

\begin{proposition}[Omission invariance]
  \label{prop:omission}
  Let $B$ denote the RGB tokens and $U$ the retained depth, semantic and action tokens. Under the attention mask of Eq.~\eqref{eq:mask} and the modality-dependent routing of Eq.~\eqref{eq:moe}, fix the shared context, retained conditions, position identifiers, initial noise of $U$ and denoising schedule. In exact arithmetic and without step caching, the retained trajectories and predicted action chunk are identical with and without $B$.
\end{proposition}

\noindent\textit{Proof sketch.}
Every query in $U$ attends only to keys in $U$. Residual connections, token-wise operations and modality-dependent routing introduce no dependence on $B$, and retained tokens keep the same position identifiers. Each retained operation therefore receives identical inputs with and without $B$. Induction over layers and denoising steps yields identical retained trajectories and action predictions.

\paragraph{Numerical execution.}
Removing tokens can alter floating-point kernels and accumulation order. The cosine step cache compares predictions across active streams, so RGB removal can also alter cache decisions. Benchmark deployment caches the retained streams. Paired consistency measurements disable caching and compilation and use identical initial noise for the retained streams.

\paragraph{Paired checkpoint evaluation.}
Tab.~\ref{tab:omission-results} compares three execution modes of the same RGB-isolated RoboTwin checkpoint on one Open Microwave observation. Retained conditions, initial noise and the ten denoising steps are identical across modes. All three produce the same normalized action chunk; complete RGB omission reduces median latency from 1,388.7 to 960.4\,ms (30.8\%). Each mode uses one warmup request and ten timed requests on a single GPU with PyTorch 2.7.1+cu128. Caching and compilation are disabled. DINO, depth and text conditions are precomputed; timing includes joint denoising and RGB condition encoding when used. World decoding is disabled, and all model parameters remain resident during peak allocated-memory measurement.

\begin{table}[H]
  \centering\footnotesize
  \caption{\textbf{RGB-generation omission at inference.} All modes use the same retained conditions and initial noise. Same action: the normalized action chunk is identical to that of Full RGB.}
  \label{tab:omission-results}
  \setlength{\tabcolsep}{10pt}
  \begin{tabular*}{\textwidth}{@{\extracolsep{\fill}}lcccc@{}}
    \toprule
    Execution mode & Latency (ms/chunk) & Speed-up & Peak memory (GB) & Same action \\
    \midrule
    Full RGB & 1388.7 & -- & 29.249 & -- \\
    Future RGB omitted & 1206.4 & 13.1\% & 29.233 & \checkmark \\
    Entire RGB omitted & \textbf{960.4} & \textbf{30.8\%} & 29.199 & \checkmark \\
    \bottomrule
  \end{tabular*}
\end{table}

\section{Evaluation Protocol}
\label{supp:eval}
The evaluation covers RoboTwin-C2R, RoboTwin-Full, LIBERO and LIBERO-Plus. This section specifies their data splits, success-rate aggregation and inference settings.

\paragraph{Benchmarks and aggregation.}
RoboTwin-C2R uses 2,450 clean training episodes and 50 validation episodes. Evaluation comprises 100 episodes per task and condition across 50 tasks. RoboTwin-Full uses 50 clean and 500 randomized demonstrations per task. RoboTwin success rates are averaged over tasks; Appendix~\ref{supp:pertask} reports individual task results. LIBERO uses 50 trials per task. LIBERO-Plus evaluates the LIBERO checkpoint without additional fine-tuning and aggregates 10,030 trials across seven perturbation dimensions.

\paragraph{Policy and scene configuration.}
Tab.~\ref{tab:train-config} lists the training and inference settings. Evaluation uses model seed 42, scene seed 42 (10000 on LIBERO-Plus), ten denoising steps, guidance scale 1, 32 executed actions per chunk (10 on LIBERO and LIBERO-Plus) and action normalization fixed from the training set. Cosine step caching uses threshold 0.99 and at most three consecutive skips; compilation is disabled. Scene and instruction assignments are fixed independently of the policy, and controlled comparisons index clutter offsets by the global episode index. Each reported result comes from a single complete evaluation run, and success rates include all evaluated episodes.

\begin{table}[H]
  \centering
  \caption{\textbf{Training and inference settings.} RoboTwin-C2R and RoboTwin-Full differ only in training data: clean demonstrations for C2R, clean and randomized demonstrations for Full. LIBERO and LIBERO-Plus use the same checkpoint.}
  \label{tab:train-config}
  \footnotesize
  \setlength{\tabcolsep}{3pt}
  \begin{tabularx}{\textwidth}{@{}l*{3}{>{\centering\arraybackslash}X}@{}}
    \toprule
    Setting & RoboTwin & LIBERO & Real world \\
    \midrule
    Epochs / global batch & 15 / 256 & 30 / 256 & 30 / 128 \\
    Learning rate & $10^{-4}$ & $10^{-4}$ & $10^{-4}$ \\
    Condition noise ($p$ / $\alpha_{\max}$) & 0.1 / 0.2 & 0.1 / 0.2 & 0.1 / 0.2 \\
    States / actions / frames & 33 / 32 / 9 & 33 / 32 / 9 & 33 / 32 / 9 \\
    Action space & 14-D joint & 7-D end-effector & end-effector + gripper \\
    \addlinespace[2pt]
    Camera views & head + two wrist & main + wrist & head + two wrist \\
    Canvas (H$\times$W) & $384\times320$ & $224\times448$ & $384\times320$ \\
    Depth source & simulator GT & simulator GT & Video Depth Anything \\
    Semantic readout & head view & main view & head view \\
    \addlinespace[2pt]
    Denoising steps / executed actions & 10 / 32 & 10 / 10 & 10 / 32 \\
    \bottomrule
  \end{tabularx}
\end{table}

\section{Baseline Sources and Input Modalities}
\label{supp:comparison}
The baseline comparisons combine published results obtained under each method's training and evaluation settings. We identify the sources for each benchmark and distinguish inference inputs from spatial supervision used during training.

\paragraph{Result sources.}
RoboTwin-C2R results in Tab.~\ref{tab:comparison-robotwin}, including PrismWAM, come from Table~1 of PrismWAM~\citep{prismwam2026}; LIBERO and LIBERO-Plus results in Tab.~\ref{tab:comparison-libero} come from its Table~2. The PrismWAM entry uses the Wan2.2-5B-based model with approximately 6B parameters.

For RoboTwin-Full, $\pi_{0.5}$ and FastWAM results come from Fast-WAM~\citep{fastwam2026}, OpenWAM-$\alpha$ and starVLA-OFT results from OpenWAM~\citep{openwam_pku2026}, and X-WAM and 4D-WAM results from their respective papers~\citep{xwam2026,fourdwam2026}. The 4D-WAM Full and C2R results use FastWAM-Joint and LingBot-VA as base models, respectively. Easy and Hard values retain the reported precision; Avg.\ is their arithmetic mean, and a dash denotes an unavailable result.

The per-task baselines in Tab.~\ref{tab:robotwin-per-task} come from the RoboTwin~2.0 leaderboard. Their means can differ slightly from the PrismWAM-reported values in Tab.~\ref{tab:comparison-robotwin}.

\paragraph{Input modalities.}
Tab.~\ref{tab:baseline-inputs} distinguishes observed geometry from geometry used solely as a training target. Language and robot-state inputs follow each method's evaluation protocol. In simulation, both \ourmethod and PrismWAM receive current ground-truth depth from the simulator.

\begin{table}[H]
  \centering\footnotesize
  \caption{\textbf{Visual inputs and spatial supervision of the compared methods.} Observed inputs are those used in simulation evaluation; spatial learning lists geometry used as a training target rather than as an input.}
  \label{tab:baseline-inputs}
  \setlength{\tabcolsep}{8pt}
  \begin{tabular*}{\textwidth}{@{\extracolsep{\fill}}lll@{}}
    \toprule
    Method & Observed visual input & Spatial learning \\
    \midrule
    $\pi_{0.5}$~\citep{pi05_2025} & RGB & VLM / action pretraining \\
    OpenWAM-$\alpha$~\citep{openwam_pku2026} & RGB & RGB world--action pretraining \\
    X-WAM~\citep{xwam2026} & RGB & Predicted depth / RGB-D reconstruction \\
    PrismWAM~\citep{prismwam2026} & RGB + current depth & Depth prediction; simulator GT \\
    starVLA~\citep{starvla2026} & RGB & Vision-language features \\
    FastWAM~\citep{fastwam2026} & RGB & RGB prediction during training \\
    Spatial-Forcing~\citep{spatialforcing2025} & RGB & Pretrained geometry teacher \\
    4D-WAM~\citep{fourdwam2026} & RGB & Trajectory-field feature alignment \\
    \midrule
    \ourmethod & RGB + current depth & DINOv3 + depth prediction; simulator GT \\
    \bottomrule
  \end{tabular*}
\end{table}

\section{Inference Latency and Memory Usage}
\label{supp:details}
We measure inference latency and peak allocated memory for \ourmethod, OpenWAM-$\alpha$ and PrismWAM on the same single-GPU hardware configuration. The comparison covers each method's default deployment preset and alternative caching and compilation settings.

\paragraph{Measurement protocol.}
Tab.~\ref{tab:inference-cost} reports median, 10th--90th percentile and mean chunk latency over 100 timed requests on five observations, together with first-call time and peak allocated memory. Per-action latency is the median chunk latency divided by the number of executed actions per chunk. OpenWAM-$\alpha$ uses PyTorch 2.7.1+cu128; PrismWAM and \ourmethod use 2.10.0+cu128.

\paragraph{Execution configurations.}
Step cache reuses predictions according to cosine similarity; Step cache + compile additionally enables compilation; No cache/compile evaluates every denoising step. The \ourmethod Default and Step cache rows are separate measurements of the same configuration, with threshold 0.99 and at most three consecutive skips. OpenWAM-$\alpha$ Step cache runs 1 and 2 are also separate measurements. All \ourmethod configurations omit RGB generation. The paired experiment in Appendix~\ref{supp:omission} measures the effect of RGB omission separately.

\paragraph{Latency comparison.}
Under the default presets, \ourmethod requires 17.94\,ms per executed action, compared with 5.58\,ms for OpenWAM-$\alpha$ and 33.26\,ms for PrismWAM. Without caching or compilation, the corresponding values are 42.19, 26.74 and 32.45\,ms.

\begin{table}[H]
  \centering
  \caption{\textbf{Single-GPU inference cost.} Chunk and per-action latency in ms, first-call time in s and peak allocated memory in GB. P10--P90 is the 10th--90th percentile range; Default is each method's deployment preset.}
  \label{tab:inference-cost}
  \footnotesize
  \setlength{\tabcolsep}{4pt}
  \begin{tabular*}{\textwidth}{@{\extracolsep{\fill}}lllcccccc@{}}
    \toprule
    Method & Horizon & Configuration & Median & P10--P90 & Mean & Per action & First call & Peak \\
    \midrule
    \multirow{5}{*}{\shortstack[l]{OpenWAM-$\alpha$\\\citep{openwam_pku2026}}} & \multirow{5}{*}{32} & Default & 178.6 & 176.5--180.2 & 178.8 & 5.58 & 67.9 & 25.2 \\
    & & Step cache (run 1) & 379.3 & 374.4--388.3 & 380.0 & 11.85 & 1.8 & 25.2 \\
    & & Step cache (run 2) & 354.3 & 351.5--364.2 & 357.7 & 11.07 & 1.7 & 25.2 \\
    & & Step cache + compile & 178.0 & 176.2--179.3 & 177.9 & 5.56 & 13.6 & 25.2 \\
    & & No cache/compile & 855.6 & 840.9--879.3 & 860.9 & 26.74 & 2.4 & 25.2 \\
    \midrule
    \multirow{4}{*}{\shortstack[l]{PrismWAM\\\citep{prismwam2026}}} & \multirow{4}{*}{24} & Default & 798.1 & 777.9--844.9 & 805.5 & 33.26 & 1.2 & 26.9 \\
    & & Step cache & 370.7 & 368.1--376.3 & 372.0 & 15.45 & 0.8 & 27.0 \\
    & & Step cache + compile & 229.7 & 227.0--237.3 & 240.2 & 9.57 & 12.1 & 27.0 \\
    & & No cache/compile & 778.8 & 771.0--834.5 & 792.6 & 32.45 & 1.2 & 26.9 \\
    \midrule
    \multirow{4}{*}{\ourmethod} & \multirow{4}{*}{32} & Default & 574.2 & 568.1--597.9 & 578.5 & 17.94 & 1.2 & 29.2 \\
    & & Step cache & 563.0 & 556.7--584.9 & 566.2 & 17.59 & 1.2 & 29.2 \\
    & & Step cache + compile & 417.1 & 411.5--424.8 & 424.3 & 13.03 & 32.3 & 29.3 \\
    & & No cache/compile & 1350.1 & 1340.6--1386.5 & 1361.7 & 42.19 & 4.2 & 29.2 \\
    \bottomrule
  \end{tabular*}
\end{table}

\section{Modality and MoE Ablation Studies}
\label{supp:ablation}
The ablations examine how modality supervision, action access and expert routing affect RoboTwin-C2R performance. All variants use the same training protocol to isolate the changed component.

\paragraph{Modality configurations.}
Tab.~\ref{tab:modality-roles} varies one modality while preserving the roles of the other two. Removed excludes the stream and its prediction loss; Auxiliary retains prediction without direct action access. Read current and Read current + future define the token groups available to the action expert. RGB-access variants retain separate RGB interfaces. The semantic head-view mask follows Appendix~\ref{supp:params}.

\paragraph{Training protocol.}
All variants train on clean RoboTwin demonstrations for 15 epochs, with global batch size 256, learning rate $10^{-4}$ and seed 42. Camera layout, augmentation, initialization, normalization and inference horizon match \ourmethod. Modality removal changes the prediction targets, while action-access comparisons retain the same prediction tasks.

\paragraph{Expert routing.}
Tab.~\ref{tab:moe-ablation} replaces the three modality-routed MoE blocks with dense FFNs while keeping all other architecture and training settings fixed. Dense MoE improves Easy and Hard success by 1.30 and 3.86 percentage points, respectively, raising their mean from 70.54\% to 73.12\%. The larger gain on Hard is consistent with a greater benefit from modality-specific capacity under distribution shift.

\begin{table}[H]
  \centering\footnotesize
  \caption{\textbf{MoE ablation on RoboTwin-C2R (success, \%).} Dense FFN replaces the three modality-routed MoE blocks.}
  \label{tab:moe-ablation}
  \setlength{\tabcolsep}{10pt}
  \begin{tabular*}{\textwidth}{@{\extracolsep{\fill}}lccc@{}}
    \toprule
    World-tower FFN & Easy & Hard & Avg. \\
    \midrule
    Dense FFN & 89.74 & 51.34 & 70.54 \\
    Dense MoE (\ourmethod) & \textbf{91.04} & \textbf{55.20} & \textbf{73.12} \\
    \bottomrule
  \end{tabular*}
\end{table}

\section{Representation Robustness and Task Execution}
\label{supp:qual}
We assess representation stability under appearance perturbations and compare task executions from matched initial scenes. The analysis combines feature visualization, task-neighborhood entropy and state retrieval on RoboTwin and LIBERO-Plus with paired RoboTwin rollouts.

\subsection{Representation Robustness}
\begin{figure}[H]
  \centering
  \includegraphics[width=0.8\linewidth]{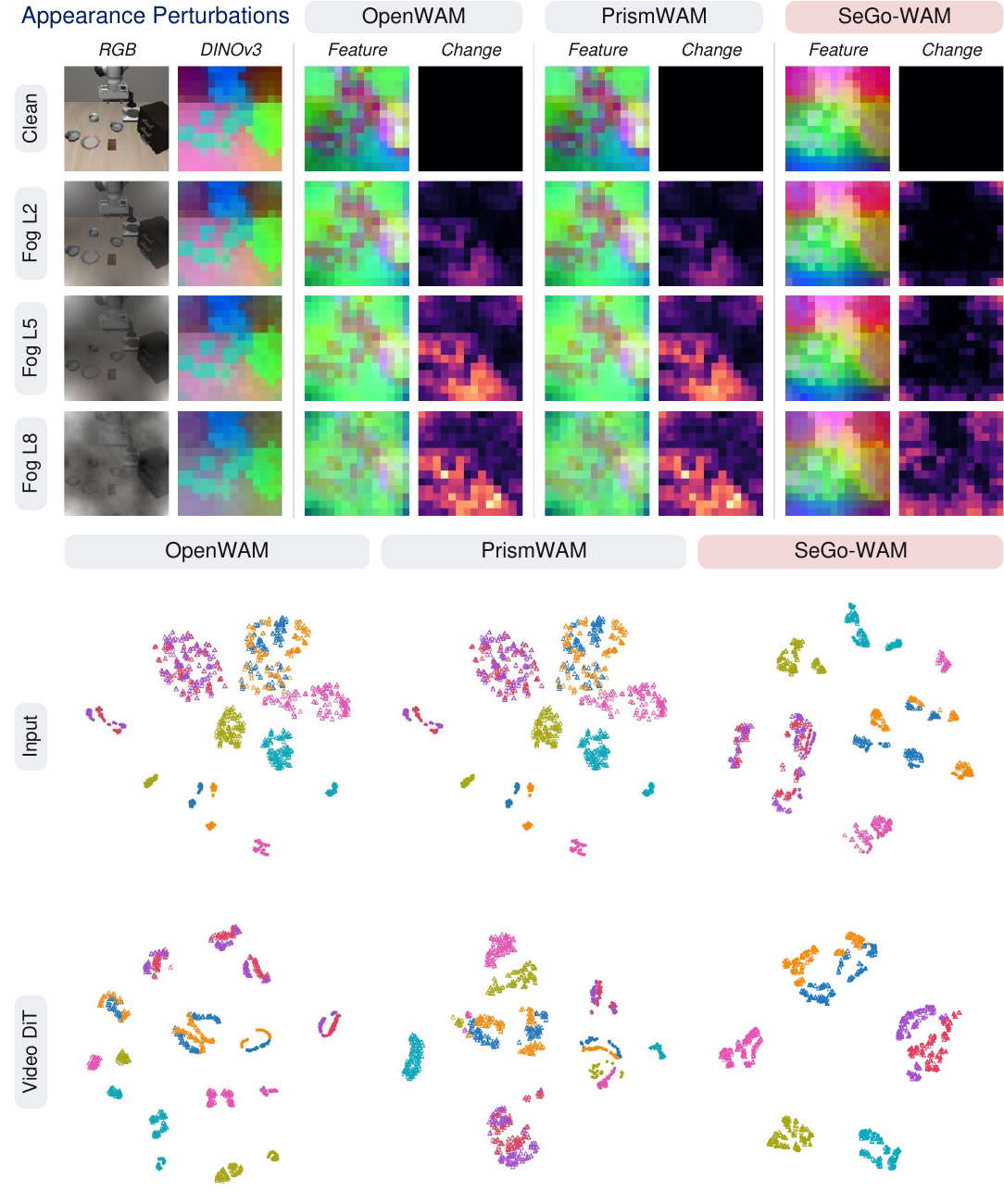}
  \caption{\textbf{Representations under LIBERO-Plus fog.} \textbf{Top}: inputs, feature maps and cosine distance to the clean features (brighter: larger change). \textbf{Bottom}: t-SNE of input encodings and final-layer features. Colors denote tasks; filled and open markers denote clean and fogged inputs.}
  \label{fig:qualitative-libero}
\end{figure}

Clean and perturbed observations of the same state allow direct comparison of input encodings and final-layer Video DiT features. RoboTwin uses lighting shifts, and LIBERO-Plus uses fog perturbations (Figs.~\ref{fig:qualitative} and~\ref{fig:qualitative-libero}).

\paragraph{Sampling.}
Each benchmark includes ten tasks, four demonstration trajectories per task and ten sampled states per trajectory, giving 400 base states. Pairing each clean state with three perturbation levels yields 1,600 observations per model and benchmark. RoboTwin uses blue-cast perturbations; LIBERO uses LIBERO-Plus fog on the main view. LIBERO wrist images, physical states and depth inputs remain fixed.

\paragraph{Feature extraction.}
Inputs share image orientation, resizing and view composition within each benchmark, followed by encoder-specific normalization. OpenWAM and PrismWAM use frozen RGB VAEs that produce matching encodings for these inputs. \ourmethod uses frozen DINOv3 features compressed to S-VAE posterior means. Video DiT features are extracted after layer 30: current RGB tokens for OpenWAM and current semantic tokens for \ourmethod at the final denoising step, and the cached RGB observation prefix for PrismWAM. RoboTwin uses each model's RoboTwin checkpoint; LIBERO-Plus uses its LIBERO checkpoint. Spatial pooling produces fixed-length descriptors.

\paragraph{Visualization.}
Each t-SNE~\citep{tsne2008} fit uses all ten tasks, seed 42 and perplexity 30. Seven tasks (1,120 points) are displayed, selected by visual inspection to reduce overlap between point clouds. The displayed subsets exclude roller grasping, mug hanging and pot lifting on RoboTwin, and stove-bowl placement, sauce placement and stove operation on LIBERO. All quantitative measurements use all ten tasks.

The feature maps show microwave opening under the strongest RoboTwin blue cast (Fig.~\ref{fig:qualitative}) and table-bowl placement under LIBERO fog levels 2, 5 and 8 (Fig.~\ref{fig:qualitative-libero}). The DINOv3 column shows normalized last-layer patch features before S-VAE compression. Feature colors use three-component PCA fitted separately to each representation's clean features. Clean and perturbed views share the 1st--99th percentile color range. Calibration uses all 400 LIBERO states and the first state of each RoboTwin task. Change maps show per-location cosine distance from the clean feature on a common 0--1 scale. In the t-SNE plots, colors identify tasks; filled and open markers denote clean and perturbed observations.

\paragraph{Quantitative measures.}
Descriptors are L2-normalized. For each observation, we find its 30 nearest neighbors by cosine distance, excluding all versions of the same base state. Mean normalized task-neighborhood entropy is
\begin{equation}
  H_{\rm task}=-\frac{1}{N\log 10}\sum_{i=1}^{N}\sum_{c=1}^{10}p_{ic}\log p_{ic},
\end{equation}
where $N$ is the number of observations and $p_{ic}$ is the fraction of observation $i$'s neighbors belonging to task $c$. Lower entropy indicates less mixing between tasks. State-retrieval R@1 measures how often a descriptor from the strongest perturbation retrieves its matching state from the 400 clean descriptors.

At the Video DiT output, \ourmethod achieves task-neighborhood entropy of 0.095 on RoboTwin and 0.123 on LIBERO-Plus, with R@1 of 99.0\% and 100.0\%, respectively. OpenWAM reaches R@1 of 14.75\% and 33.0\%, and PrismWAM 1.0\% and 46.25\% (Tab.~\ref{tab:retrieval}). At the LIBERO input, semantic features have lower R@1 than RGB features (54.5\% versus 82.25\%). The semantic retrieval advantage therefore emerges within the Video DiT.

\begin{table}[H]
  \centering\footnotesize
  \caption{\textbf{State retrieval under appearance perturbations.} R@1 (\%) of Video DiT output features against 400 clean states, under the strongest blue cast on RoboTwin and fog on LIBERO-Plus.}
  \label{tab:retrieval}
  \setlength{\tabcolsep}{10pt}
  \begin{tabular*}{\textwidth}{@{\extracolsep{\fill}}llcc@{}}
    \toprule
    Method & Action-read tokens & RoboTwin & LIBERO-Plus \\
    \midrule
    OpenWAM~\citep{openwam_pku2026} & Current RGB & 14.75 & 33.00 \\
    PrismWAM~\citep{prismwam2026} & Cached RGB prefix & 1.00 & 46.25 \\
    \ourmethod & Current semantics & \textbf{99.00} & \textbf{100.00} \\
    \bottomrule
  \end{tabular*}
\end{table}

\FloatBarrier
\clearpage
\subsection{Paired Task Execution}
We compare four selected RoboTwin executions from matched initial scenes and instructions. Both models use ten denoising steps, with execution horizons of 32 actions for \ourmethod and 24 for PrismWAM.

\begin{figure}[H]
  \centering
  \includegraphics[width=0.7\linewidth]{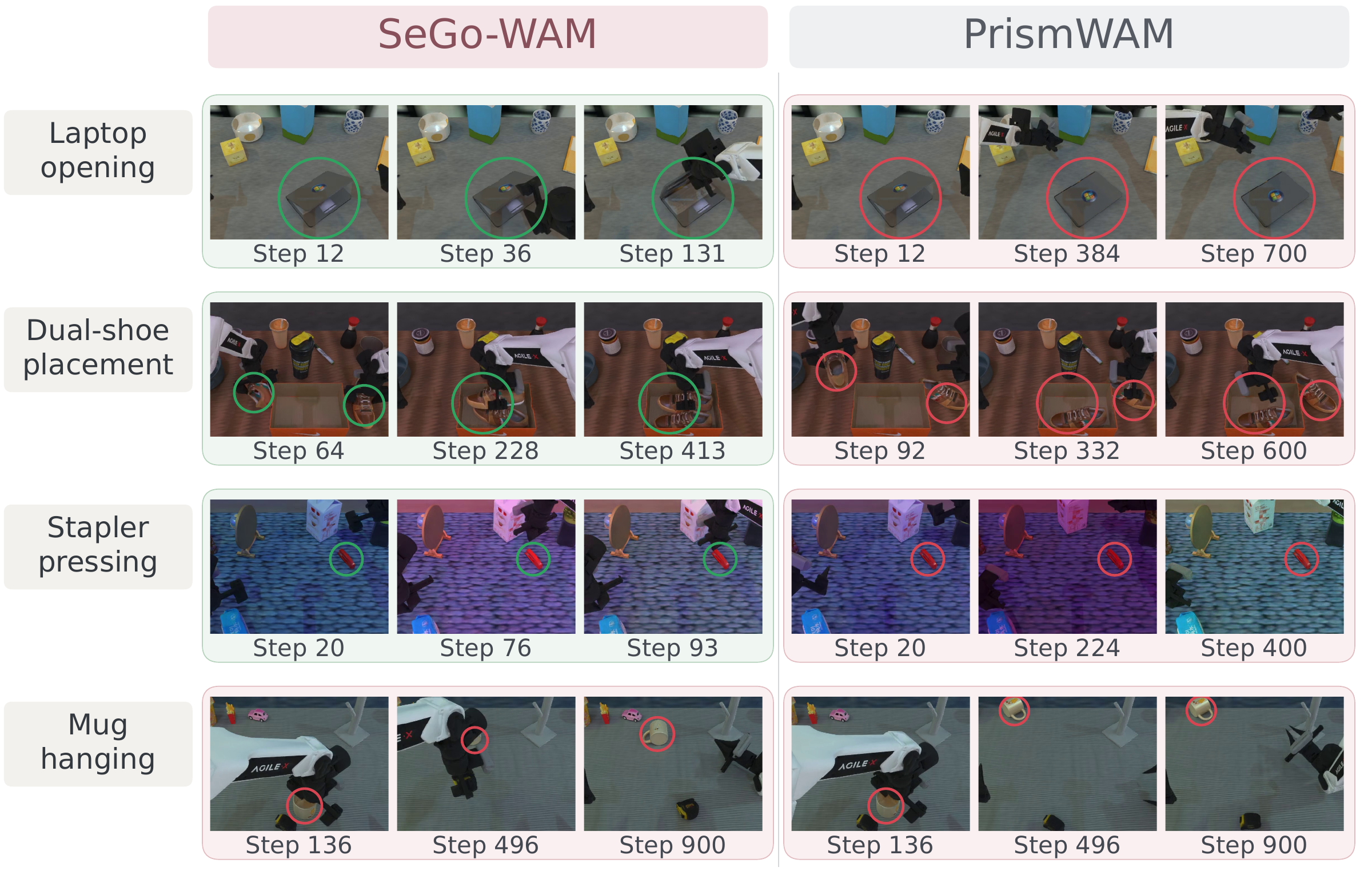}
  \caption{\textbf{Paired RoboTwin rollouts of \ourmethod (left) and PrismWAM (right).} Green and red shading denote success and failure, circles mark target objects, and frame labels give control steps.}
  \label{fig:paired-rollouts}
\end{figure}

In Fig.~\ref{fig:paired-rollouts}, \ourmethod completes laptop opening and places both shoes in the box. PrismWAM does not complete laptop opening and leaves one shoe outside the box. \ourmethod also completes stapler pressing, while PrismWAM fails to establish the required contact within the episode. Mug hanging requires grasping and precise alignment of the handle with the rack; neither model completes placement in the displayed pair.
\FloatBarrier

\section{Real-World Evaluation and Task Executions}
\label{supp:realworld}
Real-world evaluation measures sequential task completion in phone packing and examines executions of bowl stacking and test-tube transfer. We specify the demonstration data, perturbation conditions and scoring protocol for the results in Fig.~\ref{fig:realworld-task} and Tab.~\ref{tab:realworld-stages}.

\paragraph{Demonstration data.}
The phone-packing dataset contains 305 teleoperated demonstrations, split 95/5 by episode with seed 42. Each recording includes one head view, two wrist views and end-effector actions with gripper commands. Frozen Video Depth Anything~\citep{video_depth_anything} estimates depth from RGB, and the frozen depth VAE encodes these estimates.

\paragraph{Evaluation conditions.}
\ourmethod and $\pi_{0.5}$ are fine-tuned on the same demonstrations and each evaluated over 20 rollouts per condition. Clean follows the training distribution, Light changes illumination intensity and color temperature, and Object adds randomly placed distractors. A rollout receives one point for each completed sequential stage: picking up the phone, placing it in the box and closing the lid.

\paragraph{Progress metric.}
For $N$ rollouts with stage scores $q_i\in\{0,1,2,3\}$, normalized progress is
\begin{equation}
  P=\frac{100}{3N}\sum_{i=1}^{N}q_i.
\end{equation}
Tab.~\ref{tab:realworld-stages} reports total stage points and normalized progress for each condition. The perturbed average pools the 40 Light and Object rollouts, with 120 possible stage points.

\begin{table}[H]
  \centering\footnotesize
  \caption{\textbf{Real-world phone-packing progress.} Stage points per condition (maximum 60: 20 rollouts $\times$ 3 stages) and normalized progress $P$ (\%). Perturbed pools the Light and Object rollouts (maximum 120).}
  \label{tab:realworld-stages}
  \setlength{\tabcolsep}{8pt}
  \begin{tabular*}{\textwidth}{@{\extracolsep{\fill}}lcccccccc@{}}
    \toprule
    & \multicolumn{4}{c}{Stage points} & \multicolumn{4}{c}{Normalized progress $P$ (\%)} \\
    \cmidrule(lr){2-5}\cmidrule(l){6-9}
    Method & Clean & Light & Object & Perturbed & Clean & Light & Object & Perturbed \\
    \midrule
    $\pi_{0.5}$ & 35 & 25 & 30 & 55 & 58.3 & 41.7 & 50.0 & 45.8 \\
    \ourmethod & \textbf{47} & \textbf{34} & \textbf{39} & \textbf{73} & \textbf{78.3} & \textbf{56.7} & \textbf{65.0} & \textbf{60.8} \\
    \bottomrule
  \end{tabular*}
\end{table}

\begin{figure*}[p]
  \centering
  \includegraphics[width=\linewidth,height=0.92\textheight,keepaspectratio]{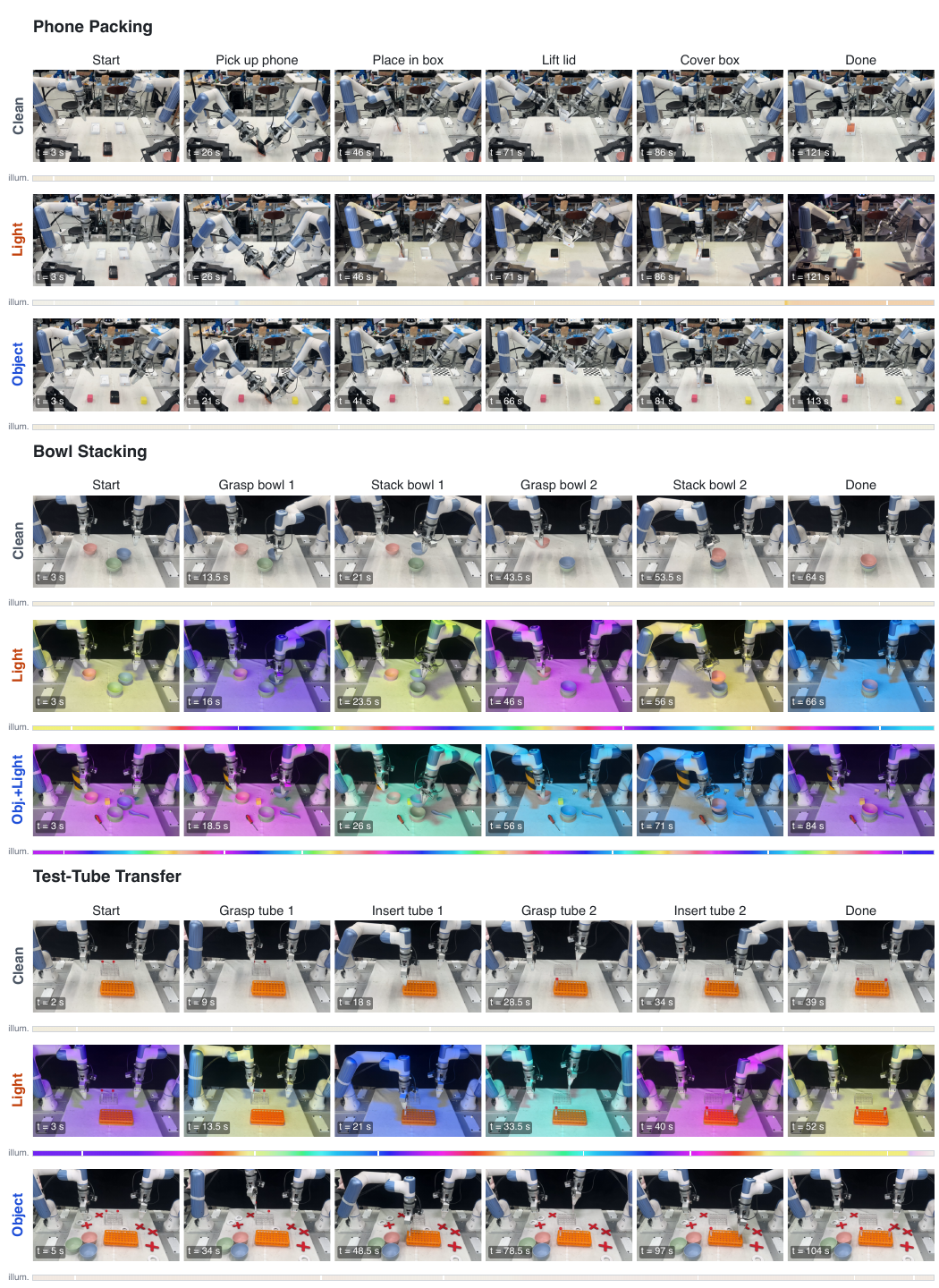}
  \caption{\textbf{Real-world executions.} Phone packing, bowl stacking and test-tube transfer, from top to bottom. Rows are evaluation conditions and columns are successive keyframes labeled with elapsed time; the bar shows the current illumination color.}
  \label{fig:realworld-strips}
\end{figure*}

\paragraph{Additional task executions.}
Fig.~\ref{fig:realworld-strips} shows complete executions of phone packing, bowl stacking and test-tube transfer. Bowl stacking requires grasping two bowls in sequence and stacking them onto a third. Test-tube transfer requires grasping two tubes from a holder and inserting each into a rack. For these two tasks, Light uses dynamic illumination that cycles through saturated colors, and Object adds distractors such as bowls, tools and printed markers. Bowl stacking also combines both perturbations in Obj.+Light. Phone packing uses fixed illumination changes. All stages are completed in the displayed executions. The additional tasks are evaluated qualitatively; quantitative progress scores are computed from phone-packing trials.

\clearpage

\section{Per-Task RoboTwin Results}
\label{supp:pertask}
Tabs.~\ref{tab:robotwin-per-task} and~\ref{tab:robotwin-full-per-task} report success rates for all 50 tasks under RoboTwin-C2R and RoboTwin-Full, respectively, complementing the aggregate results in Tab.~\ref{tab:comparison-robotwin}.

On RoboTwin-C2R, \ourmethod exceeds, matches and falls below PrismWAM on 28/9/13 Easy tasks and 30/0/20 Hard tasks, with mean gains of 3.06 and 2.04 percentage points. The largest Hard gains include Place Dual Shoes (+44 points), Handover Mic (+33), Press Stapler (+28) and Adjust Bottle (+23). The C2R comparison includes leaderboard methods with published per-task results; OpenWAM-$\alpha$ has no reported per-task values.

\begin{table}[H]
  \centering
  \caption{\textbf{Per-task success rates (\%) on RoboTwin-C2R.} Baseline results are taken from the RoboTwin~2.0 leaderboard.}
  \label{tab:robotwin-per-task}
  \footnotesize
  \setlength{\tabcolsep}{1.5pt}
  \begin{tabular*}{\textwidth}{@{\extracolsep{\fill}}l*{16}{c}@{}}
    \toprule
    & \multicolumn{2}{c}{$\pi_{0.5}$} & \multicolumn{2}{c}{starVLA} & \multicolumn{2}{c}{FastWAM} & \multicolumn{2}{c}{X-WAM} & \multicolumn{2}{c}{SF} & \multicolumn{2}{c}{4D-WAM} & \multicolumn{2}{c}{PrismWAM} & \multicolumn{2}{c}{\ourmethod} \\
    \cmidrule(lr){2-3}\cmidrule(lr){4-5}\cmidrule(lr){6-7}\cmidrule(lr){8-9}\cmidrule(lr){10-11}\cmidrule(lr){12-13}\cmidrule(lr){14-15}\cmidrule(lr){16-17}
    Task & Easy & Hard & Easy & Hard & Easy & Hard & Easy & Hard & Easy & Hard & Easy & Hard & Easy & Hard & Easy & Hard \\
    \midrule
    Adjust Bottle & 99 & \textbf{77} & 84 & 0 & 99 & 0 & 92 & 13 & 97 & 71 & 80 & 36 & 98 & 54 & \textbf{100} & \textbf{77} \\
    Beat Block Hammer & 93 & 23 & 75 & 2 & 82 & 0 & 92 & 21 & 74 & 13 & 92 & 28 & 81 & \textbf{48} & \textbf{96} & 40 \\
    Blocks Ranking RGB & 72 & \textbf{44} & 39 & 2 & 81 & 0 & 85 & 36 & 83 & 7 & 88 & 10 & 87 & 32 & \textbf{91} & 38 \\
    Blocks Ranking Size & 45 & 21 & 14 & 0 & 52 & 0 & 33 & 0 & 42 & 3 & 54 & 8 & 56 & 27 & \textbf{69} & \textbf{49} \\
    Click Alarmclock & 65 & 50 & 60 & 0 & 99 & 60 & 95 & 61 & 89 & 28 & \textbf{100} & 70 & \textbf{100} & \textbf{86} & \textbf{100} & 73 \\
    \addlinespace[2pt]
    Click Bell & 31 & 35 & 75 & 7 & 95 & 7 & \textbf{100} & 52 & 97 & 28 & 98 & 77 & \textbf{100} & \textbf{78} & \textbf{100} & 68 \\
    Dump Bin Bigbin & 92 & \textbf{84} & 54 & 4 & \textbf{94} & 0 & 88 & 40 & 83 & 70 & 86 & 57 & 84 & 80 & 93 & 83 \\
    Grab Roller & 99 & 97 & 94 & 44 & \textbf{100} & 2 & 95 & 16 & 98 & 94 & \textbf{100} & 68 & \textbf{100} & 89 & \textbf{100} & \textbf{100} \\
    Handover Block & 30 & 12 & 6 & 0 & 66 & 0 & 21 & 21 & 62 & 2 & 79 & 25 & \textbf{99} & \textbf{37} & \textbf{99} & 10 \\
    Handover Mic & 98 & 6 & 66 & 1 & 98 & 0 & 97 & 8 & \textbf{100} & 2 & 85 & 13 & 95 & 24 & 98 & \textbf{57} \\
    \addlinespace[2pt]
    Hanging Mug & 17 & 14 & 1 & 0 & 14 & 0 & 17 & 0 & 17 & 1 & \textbf{70} & 10 & 38 & 17 & 68 & \textbf{26} \\
    Lift Pot & 98 & 35 & 68 & 0 & 91 & 0 & 77 & 0 & 98 & 0 & 88 & 9 & 99 & \textbf{67} & \textbf{100} & 14 \\
    Move Can Pot & 60 & 10 & 33 & 0 & 96 & 0 & 77 & \textbf{34} & 69 & 1 & 94 & 17 & \textbf{99} & 19 & 98 & 15 \\
    Move Pillbottle Pad & 46 & \textbf{32} & 40 & 0 & 92 & 0 & 78 & 26 & 77 & 14 & 89 & 22 & \textbf{100} & 19 & \textbf{100} & 27 \\
    Move Playingcard Away & 84 & \textbf{65} & 62 & 0 & 94 & 0 & 83 & 11 & 96 & 15 & 82 & 35 & 99 & \textbf{65} & \textbf{100} & 46 \\
    \addlinespace[2pt]
    Move Stapler Pad & 20 & \textbf{18} & 14 & 0 & 33 & 0 & 26 & 2 & 31 & 1 & 43 & 12 & 33 & 8 & \textbf{44} & 9 \\
    Open Laptop & 93 & 68 & 22 & 2 & 84 & 0 & 72 & 6 & 89 & 15 & 85 & \textbf{82} & 91 & 50 & \textbf{99} & 58 \\
    Open Microwave & \textbf{86} & \textbf{59} & 10 & 1 & 17 & 0 & 9 & 24 & 66 & 18 & 68 & 12 & 57 & 19 & 29 & 23 \\
    Pick Diverse Bottles & 66 & 42 & 43 & 15 & 72 & 0 & 74 & 26 & 72 & 38 & 77 & 44 & \textbf{93} & 78 & 88 & \textbf{81} \\
    Pick Dual Bottles & 72 & 57 & 60 & 20 & 87 & 1 & 99 & 55 & 90 & 47 & 92 & 67 & \textbf{100} & \textbf{96} & \textbf{100} & 88 \\
    \addlinespace[2pt]
    Place A2B Left & 71 & \textbf{49} & 24 & 0 & 69 & 0 & 70 & 18 & 67 & 11 & 71 & 19 & 83 & 39 & \textbf{95} & 38 \\
    Place A2B Right & 68 & 39 & 25 & 1 & 60 & 1 & 59 & 21 & 55 & 6 & 62 & 42 & 80 & 41 & \textbf{92} & \textbf{52} \\
    Place Bread Basket & 63 & 60 & 82 & 5 & 87 & 0 & 82 & 40 & 84 & 53 & 87 & 77 & 94 & 77 & \textbf{96} & \textbf{81} \\
    Place Bread Skillet & 76 & 47 & 79 & 2 & 85 & 0 & 82 & 37 & 82 & 30 & 82 & 69 & 90 & 66 & \textbf{93} & \textbf{75} \\
    Place Burger Fries & 82 & 86 & 85 & 11 & 97 & 0 & 93 & 36 & \textbf{98} & 71 & 86 & 72 & \textbf{98} & 90 & \textbf{98} & \textbf{97} \\
    \addlinespace[2pt]
    Place Can Basket & 56 & 10 & 4 & 0 & 77 & 0 & 63 & 15 & 77 & 10 & \textbf{83} & \textbf{51} & 81 & 31 & 80 & 35 \\
    Place Cans Plasticbox & 31 & 73 & 95 & 2 & 98 & 0 & 93 & 36 & 94 & 11 & 94 & 48 & \textbf{100} & 95 & 99 & \textbf{100} \\
    Place Container Plate & 92 & 72 & 87 & 3 & \textbf{100} & 0 & 95 & 52 & 97 & 54 & 93 & \textbf{82} & \textbf{100} & 68 & 96 & 81 \\
    Place Dual Shoes & 70 & 45 & 30 & 2 & 67 & 0 & 38 & 7 & 83 & 2 & 66 & 19 & 73 & 43 & \textbf{90} & \textbf{87} \\
    Place Empty Cup & 96 & 80 & 85 & 0 & 90 & 0 & 93 & 36 & 94 & 33 & 99 & 70 & 99 & 70 & \textbf{100} & \textbf{90} \\
    \addlinespace[2pt]
    Place Fan & 66 & 26 & 30 & 1 & 59 & 0 & 47 & 6 & 74 & 7 & 48 & 17 & 88 & \textbf{42} & \textbf{96} & 31 \\
    Place Mouse Pad & 34 & \textbf{31} & 11 & 0 & 55 & 0 & 49 & 6 & 35 & 9 & 54 & 9 & \textbf{89} & 29 & 87 & 17 \\
    Place Object Basket & 70 & 29 & 4 & 1 & 82 & 0 & 52 & 20 & 79 & 16 & 75 & 40 & 86 & \textbf{44} & \textbf{89} & 42 \\
    Place Object Scale & 78 & 38 & 20 & 1 & 83 & 0 & 66 & 14 & 73 & 4 & 79 & 24 & 88 & 37 & \textbf{94} & \textbf{52} \\
    Place Object Stand & 85 & 47 & 66 & 0 & 94 & 0 & 83 & 28 & 96 & 32 & 87 & 42 & 97 & 76 & \textbf{100} & \textbf{82} \\
    \addlinespace[2pt]
    Place Phone Stand & 65 & 37 & 18 & 0 & 85 & 0 & 35 & 4 & 83 & 1 & 84 & 21 & 88 & \textbf{47} & \textbf{95} & 14 \\
    Place Shoe & 82 & 50 & 46 & 0 & 87 & 0 & 76 & 31 & 88 & \textbf{61} & 83 & 52 & \textbf{92} & 55 & 90 & 40 \\
    Press Stapler & 79 & 63 & 91 & 10 & 90 & 3 & 85 & 20 & 89 & 55 & 76 & 61 & \textbf{92} & 50 & 87 & \textbf{78} \\
    Put Bottles Dustbin & 78 & 50 & 11 & 0 & 55 & 0 & 71 & 21 & 58 & 41 & 68 & 19 & \textbf{90} & 82 & \textbf{90} & \textbf{87} \\
    Put Object Cabinet & 39 & 22 & 29 & 0 & 42 & 0 & 28 & 16 & 37 & 1 & 69 & \textbf{47} & 89 & 28 & \textbf{91} & 16 \\
    \addlinespace[2pt]
    Rotate QRcode & 90 & 21 & 11 & 0 & 81 & 0 & 47 & 2 & 87 & 22 & 79 & \textbf{67} & 83 & 26 & \textbf{94} & 9 \\
    Scan Object & 49 & 40 & 63 & 5 & 68 & 0 & 32 & 14 & 76 & 6 & 73 & 53 & \textbf{94} & \textbf{72} & 93 & 65 \\
    Shake Bottle & \textbf{100} & \textbf{100} & \textbf{100} & 5 & \textbf{100} & 10 & 96 & 67 & 99 & 84 & 98 & 72 & \textbf{100} & 79 & 99 & 97 \\
    Shake Bottle Horizontally & 99 & \textbf{100} & 98 & 9 & \textbf{100} & 9 & 97 & 66 & 99 & 87 & 98 & 70 & \textbf{100} & 80 & 99 & 94 \\
    Stack Blocks Three & 77 & 25 & 20 & 1 & 89 & 0 & 72 & 26 & 75 & 0 & 95 & \textbf{82} & 90 & 33 & \textbf{98} & 28 \\
    \addlinespace[2pt]
    Stack Blocks Two & 91 & 43 & 53 & 1 & 99 & 0 & 97 & \textbf{76} & 96 & 19 & \textbf{100} & 34 & 98 & 60 & \textbf{100} & 69 \\
    Stack Bowls Three & 81 & 52 & 46 & 0 & 80 & 0 & 69 & 38 & 79 & 55 & 82 & 27 & \textbf{84} & 59 & \textbf{84} & \textbf{79} \\
    Stack Bowls Two & 94 & 70 & 65 & 0 & 94 & 0 & 93 & 67 & 96 & 79 & 94 & 81 & 94 & 74 & \textbf{97} & \textbf{85} \\
    Stamp Seal & 53 & 21 & 12 & 0 & 30 & 0 & 58 & 16 & 39 & 2 & 74 & 10 & 80 & \textbf{23} & \textbf{95} & 16 \\
    Turn Switch & 52 & 25 & 16 & 0 & 39 & 3 & 68 & 4 & 40 & 7 & \textbf{86} & 11 & 70 & \textbf{49} & 63 & 41 \\
    \midrule
    Mean (50 tasks) & 70.66 & 46.00 & 46.52 & 3.16 & 77.76 & 1.92 & 69.98 & 25.84 & 77.18 & 26.74 & 81.50 & 41.80 & 87.98 & 53.16 & \textbf{91.04} & \textbf{55.20} \\
    \bottomrule
  \end{tabular*}
  \par\smallskip\begin{minipage}{\textwidth}\footnotesize\textit{Note.} Baselines: $\pi_{0.5}$~\citep{pi05_2025}, starVLA~\citep{starvla2026}, FastWAM~\citep{fastwam2026}, X-WAM~\citep{xwam2026}, Spatial-Forcing (SF)~\citep{spatialforcing2025}, 4D-WAM~\citep{fourdwam2026} and PrismWAM~\citep{prismwam2026}. \textbf{Bold}: best per task and condition.\end{minipage}
\end{table}

\begin{table}[H]
  \centering
  \caption{\textbf{Per-task success rates (\%) on RoboTwin-Full.} Easy and Hard denote clean and randomized evaluation scenes.}
  \label{tab:robotwin-full-per-task}
  \footnotesize
  \setlength{\tabcolsep}{3pt}
  \begin{tabularx}{\textwidth}{@{}l*{8}{>{\centering\arraybackslash}X}@{}}
    \toprule
    & \multicolumn{2}{c}{$\pi_{0.5}$} & \multicolumn{2}{c}{FastWAM} & \multicolumn{2}{c}{X-WAM} & \multicolumn{2}{c}{\ourmethod} \\
    \cmidrule(lr){2-3}\cmidrule(lr){4-5}\cmidrule(lr){6-7}\cmidrule(lr){8-9}
    Task & Easy & Hard & Easy & Hard & Easy & Hard & Easy & Hard \\
    \midrule
    Adjust Bottle & \textbf{100} & 99 & \textbf{100} & \textbf{100} & \textbf{100} & 99 & \textbf{100} & \textbf{100} \\
    Beat Block Hammer & 96 & 93 & \textbf{99} & \textbf{97} & 98 & 96 & \textbf{99} & 96 \\
    Blocks Ranking RGB & 92 & 85 & \textbf{100} & \textbf{100} & 99 & 95 & 95 & 97 \\
    Blocks Ranking Size & 49 & 26 & 94 & \textbf{98} & 76 & 82 & \textbf{97} & 84 \\
    Click Alarmclock & 98 & 89 & \textbf{100} & \textbf{100} & 98 & 99 & 94 & 95 \\
    \addlinespace[2pt]
    Click Bell & 99 & 66 & \textbf{100} & \textbf{100} & \textbf{100} & \textbf{100} & \textbf{100} & \textbf{100} \\
    Dump Bin Bigbin & 92 & \textbf{97} & \textbf{97} & 96 & 90 & 96 & \textbf{97} & 95 \\
    Grab Roller & \textbf{100} & \textbf{100} & \textbf{100} & \textbf{100} & \textbf{100} & \textbf{100} & \textbf{100} & \textbf{100} \\
    Handover Block & 66 & 57 & 95 & 81 & 88 & 79 & \textbf{98} & \textbf{93} \\
    Handover Mic & 98 & 97 & \textbf{99} & \textbf{100} & 88 & 89 & 97 & 99 \\
    \addlinespace[2pt]
    Hanging Mug & 18 & 17 & \textbf{58} & \textbf{62} & 46 & 55 & 57 & 60 \\
    Lift Pot & 96 & 85 & \textbf{100} & \textbf{100} & \textbf{100} & 99 & \textbf{100} & \textbf{100} \\
    Move Can Pot & 51 & 55 & 90 & 88 & 80 & 84 & \textbf{96} & \textbf{100} \\
    Move Pillbottle Pad & 84 & 61 & \textbf{100} & 99 & 96 & 98 & \textbf{100} & \textbf{100} \\
    Move Playingcard Away & 96 & 84 & \textbf{100} & \textbf{100} & \textbf{100} & 98 & \textbf{100} & \textbf{100} \\
    \addlinespace[2pt]
    Move Stapler Pad & 56 & 42 & 77 & 64 & 67 & 70 & \textbf{84} & \textbf{90} \\
    Open Laptop & 90 & 96 & 98 & \textbf{100} & 96 & 97 & \textbf{100} & \textbf{100} \\
    Open Microwave & 34 & 77 & 62 & 45 & \textbf{89} & \textbf{92} & 55 & 47 \\
    Pick Diverse Bottles & 81 & 71 & 80 & 85 & \textbf{91} & \textbf{92} & \textbf{91} & 91 \\
    Pick Dual Bottles & 93 & 63 & \textbf{100} & 96 & 99 & \textbf{100} & \textbf{100} & \textbf{100} \\
    \addlinespace[2pt]
    Place A2B Left & 87 & 82 & 95 & 93 & 90 & 87 & \textbf{99} & \textbf{99} \\
    Place A2B Right & 87 & 84 & 93 & \textbf{99} & 92 & 89 & \textbf{100} & \textbf{99} \\
    Place Bread Basket & 77 & 64 & 91 & 93 & 90 & 91 & \textbf{96} & \textbf{97} \\
    Place Bread Skillet & 85 & 66 & 90 & 93 & 90 & \textbf{96} & \textbf{98} & 94 \\
    Place Burger Fries & 94 & 87 & 96 & \textbf{99} & \textbf{97} & \textbf{99} & \textbf{97} & \textbf{99} \\
    \addlinespace[2pt]
    Place Can Basket & 62 & 62 & 71 & 69 & \textbf{84} & \textbf{82} & 79 & 80 \\
    Place Cans Plasticbox & 94 & 84 & 99 & 96 & 99 & 98 & \textbf{100} & \textbf{100} \\
    Place Container Plate & 99 & 95 & 96 & \textbf{100} & 98 & \textbf{100} & \textbf{100} & \textbf{100} \\
    Place Dual Shoes & 75 & 75 & \textbf{94} & 88 & 83 & 81 & 92 & \textbf{89} \\
    Place Empty Cup & \textbf{100} & 99 & \textbf{100} & \textbf{100} & 98 & 99 & \textbf{100} & \textbf{100} \\
    \addlinespace[2pt]
    Place Fan & 87 & 85 & 96 & \textbf{96} & 84 & 92 & \textbf{100} & 95 \\
    Place Mouse Pad & 60 & 39 & 83 & \textbf{89} & 84 & 86 & \textbf{85} & 87 \\
    Place Object Basket & 80 & 76 & 89 & 88 & 85 & 87 & \textbf{90} & \textbf{92} \\
    Place Object Scale & 86 & 80 & 90 & \textbf{97} & 93 & 89 & \textbf{94} & \textbf{97} \\
    Place Object Stand & 91 & 85 & 90 & 94 & \textbf{97} & 96 & 96 & \textbf{98} \\
    \addlinespace[2pt]
    Place Phone Stand & 81 & 81 & 97 & \textbf{99} & 75 & 80 & \textbf{98} & \textbf{99} \\
    Place Shoe & 92 & 93 & 96 & 99 & 97 & 99 & \textbf{100} & \textbf{100} \\
    Press Stapler & 87 & 83 & 90 & \textbf{97} & 94 & 90 & \textbf{98} & 94 \\
    Put Bottles Dustbin & 84 & 79 & 95 & 90 & 85 & \textbf{95} & \textbf{96} & \textbf{95} \\
    Put Object Cabinet & 80 & 79 & \textbf{94} & \textbf{89} & 66 & 76 & 91 & 80 \\
    \addlinespace[2pt]
    Rotate QRcode & 89 & 87 & 93 & \textbf{89} & 84 & 83 & \textbf{97} & 88 \\
    Scan Object & 72 & 65 & 89 & \textbf{92} & 86 & 79 & \textbf{97} & \textbf{92} \\
    Shake Bottle & 99 & 97 & \textbf{100} & \textbf{100} & 99 & 99 & \textbf{100} & \textbf{100} \\
    Shake Bottle Horizontally & 99 & 99 & \textbf{100} & \textbf{100} & \textbf{100} & 99 & \textbf{100} & \textbf{100} \\
    Stack Blocks Three & 91 & 76 & 95 & 97 & 97 & 95 & \textbf{100} & \textbf{98} \\
    \addlinespace[2pt]
    Stack Blocks Two & 97 & \textbf{100} & \textbf{100} & \textbf{100} & \textbf{100} & \textbf{100} & \textbf{100} & \textbf{100} \\
    Stack Bowls Three & 77 & 71 & 80 & 81 & 88 & 82 & \textbf{90} & \textbf{88} \\
    Stack Bowls Two & 95 & 96 & 92 & \textbf{98} & 98 & \textbf{98} & \textbf{99} & 94 \\
    Stamp Seal & 79 & 55 & 90 & 94 & 93 & \textbf{95} & \textbf{100} & \textbf{95} \\
    Turn Switch & 62 & 54 & 61 & 59 & 61 & 72 & \textbf{75} & \textbf{83} \\
    \midrule
    Mean (50 tasks) & 82.74 & 76.76 & 91.88 & 91.78 & 89.80 & 90.70 & \textbf{94.54} & \textbf{93.58} \\
    \bottomrule
  \end{tabularx}
  \par\smallskip
  \begin{minipage}{\textwidth}
    \footnotesize\raggedright\textit{Note.} Published per-task baselines only. Sources: FastWAM~\citep{fastwam2026} for $\pi_{0.5}$ and FastWAM; the original X-WAM paper~\citep{xwam2026} for X-WAM. X-WAM means retain reported precision. \textbf{Bold}: best per task and condition.
  \end{minipage}
\end{table}

\clearpage

\end{document}